\documentclass[manuscript, screen]{acmart}

\AtBeginDocument{%
  }

\setcopyright{cc}
\setcctype{by}
\acmJournal{TIST}
\acmYear{2025} \acmVolume{16} \acmNumber{6} \acmArticle{124} \acmMonth{10} \acmPrice{}\acmDOI{10.1145/3763002}

\usepackage{multirow} 

\begin{document}

\title{Advances in Pre-trained Language Models for Domain-Specific Text Classification: A Systematic Review}

\author{Zhyar Rzgar K Rostam}
\orcid{0000-0001-9906-6883}
\affiliation{%
  \institution{Doctoral School of Applied Informatics and Applied Mathematics, Obuda University}
  \city{Budapest}
  \country{Hungary}}
\email{zhyar.rostam@stud.uni-obuda.hu}

\author{Gábor Kertész}
\orcid{0000-0002-8845-8301}
\affiliation{%
  \institution{John von Neumann Faculty of Informatics, Obuda University}
  \city{Budapest}
  \country{Hungary}}
  \affiliation{%
    \institution{Laboratory of Parallel and Distributed Systems, Institute for Computer Science and Control (SZTAKI), Hungarian Research Network (HUN-REN)}
    \city{Budapest}
    \country{Hungary}
}

\renewcommand{\shortauthors}{Rostam, Kertész}

\begin{abstract}
The exponential increase in scientific literature and online information necessitates efficient methods for extracting knowledge from textual data. Natural Language Processing (NLP) plays a crucial role in addressing this challenge, particularly in text classification tasks. While Large Language Models (LLMs) have achieved remarkable success in NLP, their accuracy can suffer in domain-specific contexts due to specialized vocabulary, unique grammatical structures, and imbalanced data distributions. In this Systematic Literature Review (SLR), we investigate the utilization of Pre-trained Language Models (PLMs) for domain-specific text classification. We systematically review 41 articles published between 2018 and January 2024, adhering to the PRISMA statement (Preferred Reporting Items for Systematic Reviews and Meta-Analyses).
This review methodology involved rigorous inclusion criteria and a multi-step selection process employing AI-powered tools. We delve into the evolution of text classification techniques and differentiate between traditional and modern approaches. We emphasize transformer-based models and explore the challenges and considerations associated with using LLMs for domain-specific text classification. Furthermore, we categorize existing research based on various PLMs and propose a taxonomy of techniques used in the field. To validate our findings, we conducted a comparative experiment involving BERT, SciBERT, and BioBERT in biomedical sentence classification. Finally, we present a comparative study on the performance of LLMs in text classification tasks across different domains. In addition, we examine recent advancements in PLMs for domain-specific text classification and offer insights into future directions and limitations in this rapidly evolving domain.
\end{abstract}

\ccsdesc[500]{General and reference~Surveys and overviews}
\ccsdesc[500]{Computing methodologies~Natural language processing}
\ccsdesc[500]{Information systems~Language models}


\keywords{Large language model (LLM), pre-trained language model (PLM), prompt-based learning, scientific text, text classification, Transformer}

\received{  2024}
\received[revised]{  2025}
\received[accepted]{ 2025}

\maketitle

\section{Introduction}
\label{sec:introduction}
In recent years, there has been rapid growth in the volume of scientific literature published in all disciplines and a vast amount of information available online~\cite{beltagy2019scibert, levine2019sensebert, gan2023sensitivity, ahanger2022novel, luo2023exploring}.  This exponential growth has intensified the demand for efficient and effective methods to extract knowledge from textual context. Natural language processing (NLP) is an essential tool for addressing this challenge, specifically in the field of text classification~\cite{beltagy2019scibert, chae2023large, abriefsurvery}. Text classification plays a crucial role in various NLP tasks, including sentiment analysis~\cite{alimova2021cross, araci2019finbert, laki2023sentiment}, topic modeling~\cite{kherwa2019topic, lezama2023integrating}, information retrieval, and natural language inference (NLI). NLI specifically involves determining whether a given statement (hypothesis) logically follows from another statement (premise), categorized as entailment, contradiction, or neutral~\cite{abriefsurvery}.  In order to handle and manage these vast amounts of data efficiently, a powerful deep learning (DL) framework is required. Large language models (LLMs), the most popular DL architecture, have achieved remarkable success in almost all NLP tasks, including text classification. However, while most LLMs are first trained on a general corpus and then fine-tuned for specific tasks, in some cases, these models do not perform as well as expected, especially when fine-tuned for domain-specific text. Domain-specific text often has specialized vocabulary, different grammatical structures, and imbalanced data distributions, which can make it challenging for LLMs to perform well~\cite{beltagy2019scibert, lee2020biobert, kim2023medibiodeberta}. This systematic literature review (SLR) aims to address the mentioned challenges and outline future research directions. The study aims and contributions are outlined below.

\noindent\emph{Aims:}
\begin{itemize}
    \item Investigate the utilization of PLMs for domain-specific text classification.
    \item Provide a comprehensive analysis of advancements in text classification techniques, with a focus on domain-specific contexts.
    \item Differentiate between traditional and modern text classification approaches, emphasizing transformer-based models.
    \item Categorize existing research on PLMs and propose a taxonomy of techniques used in the field.
    \item Present a comparative study on the performance of LLMs in text classification tasks across different domains.
    \item Conduct experiments to analyze and visualize the performance of general-purpose (BERT) and domain-specific (SciBERT, BioBERT) PLMs in biomedical sentence classification.  
    \item Identify promising areas for future research in domain-specific text classification.
\end{itemize}
\emph{Contributions:}
\begin{itemize}
    \item SLR on PLMs for Domain-Specific Text Classification: In this study we systematically review 41 articles published between 2018 and January 2024, focusing on the use of PLMs for domain-specific text classification. We follow the PRISMA statement to ensure rigorous and transparent reporting of the review process.
    \item Modern and Traditional Approaches: We provide a comparison between traditional and modern text classification techniques, with a special emphasis on transformer-based models. This helps to illustrate the evolution of text classification methods over time.
    \item Taxonomy of Techniques: We categorize existing research based on various PLMs, offering a clear taxonomy of techniques used in domain-specific text classification. This serves as a valuable resource for researchers looking to understand the landscape of methods and their applications.
    \item Performance Analysis Across Domains: We present the achieved results by LLMs in text classification tasks across different domains, such as biomedical, finance, nuclear, humanitarian, social science, and materials science. This helps to identify the strengths and weaknesses of various models in specific contexts.
    \item Challenges and Considerations: We discuss the challenges associated with using LLMs for domain-specific text classification, such as the need for extensive domain-specific training data, high computational costs, and issues related to specialized vocabulary and unique grammatical structures.
    \item Key Findings and Challenges in Various Domains: We state key findings and challenges for all core categories in both domain-specific techniques and domain-specific PLMs.
    \item Insights into Model Adaptation and Efficiency: We explore advanced techniques like transfer learning, activation fine-tuning, prompt-based learning, and their effectiveness in enhancing model performance. 
    \item Case Study: We perform an empirical comparison of BERT, BioBERT and SciBERT on a sentence classification task in the biomedical domain to explore the effectiveness of domain-specific pre-training.
    \item Future Directions: We suggest future research directions, including the development of more efficient domain adaptation techniques, optimization of computational resources, creation of high-quality domain-specific datasets, and addressing ethical concerns related to model bias and transparency.

\end{itemize}

\begin{figure}[h]
  \centering
  \includegraphics[width=\linewidth]{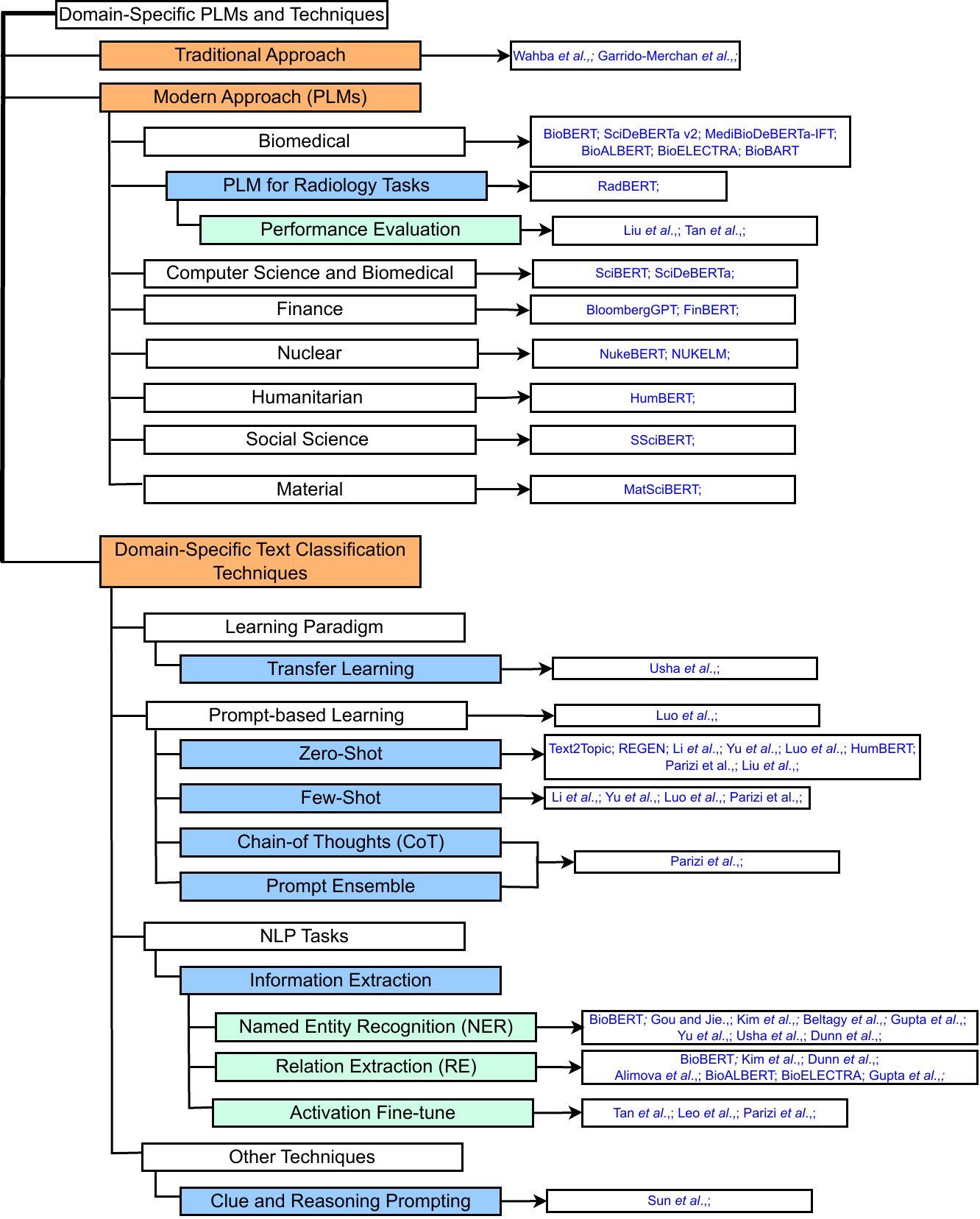}
  \caption{An overview of this SLR}
  \Description[Domain-Specific Text Classification]{This figure provides an overview of pre-trained language models (PLMs) and key techniques used in domain-specific text classification.}
  \label{fig1}
\end{figure}

In this SLR, firstly, we provide a taxonomy for PLMs and techniques utilized for domain-specific tasks (SLR overview presented in Fig.~\ref{fig1}). Then, we examine the development of text classification techniques, highlighting both traditional and recent research contributions in the field of language models (Section~\ref{sec:modern_and_traditional_text_classification_approach}). We also delve into the challenges and considerations of using LLMs for text classification (Section~\ref{subsec:domain_specific_text_classification_challenges_and_consideration}), with a specific focus on PLMs for domain-specific contexts (Section~\ref{subsec:pre_trained_language_models_for_domain_specific_text_classification}). Furthermore, we discuss the different techniques utilized for domain-specific text classification tasks (Section~\ref{sec:domain_specific_text_classification_techniques}). Moreover, we present a comparative study employing LLMs in text classification (Section~\ref{subsec:comparative_studies_for_text_classification_using_llms}). Additionally, we conduct experiments to analyze and visualize the performance of general-purpose (BERT) and domain-specific (SciBERT, BioBERT) PLMs in biomedical sentence classification (Section~\ref{expermints}). Finally, to guide future research, we outline potential future directions and acknowledge the limitations of this study (Sections~\ref{sec:future_directions} and~\ref{sec:limitation}).

\subsection{Related Works}
\label{subsec:related_works}
In this section, we will explore the most related studies conducted on pre-trained models (PTMs), and LLMs with keeping emphasis on domain-specific tasks. Jiao~\cite{abriefsurvery} offers an overview of recent advancements in text classification, delving into diverse techniques for categorizing text data. The study covers common pre-processing procedures, techniques for feature engineering, and different approaches to classification, including both machine learning and DL methods. Additionally, the study highlights the evolution of PTMs on large text corpora and their applications in text classification tasks.

Zhao et al.~\cite{zhao2023comprehensive} conduct a survey to investigate the application of deep learning for relation extraction (RE), aiming to benefit both researchers and practitioners. The study begins with fundamental elements like RE datasets and evaluation techniques, subsequently proposing a novel way to classify existing RE approaches based on text representation, context encoding, and triplet prediction. Furthermore, it systematically summarizes model architectures used in deep learning-based RE approaches and discusses challenges and solutions in diverse settings and domains, including low-resource and cross-sentence scenarios, as well as specific fields like biomedical, finance, legal, and scientific areas. 

Zhao et al.~\cite{zhao2023survey} provide an overview of recent developments in LLMs, covering key areas such as pre-training, adaptation tuning, utilization, and capacity evaluation. It serves as a valuable resource for researchers and engineers, offering insights into the latest techniques and discoveries essential for the effectiveness of LLMs. Additionally, the survey highlights available development resources and identifies remaining challenges, providing direction for future research in the field of LLMs.

Chang and Bergen~\cite{chang2024language}, comprehensively investigate more than 250 recent studies and delve into the behavior of large English language models before undergoing task-specific fine-tuning. They offer an extensive analysis of the strengths and weaknesses of the language model, aiming to be a valuable resource. While advancements in model size have significantly improved text quality for practical applications and related research, the models still struggle with generating factual information, making common-sense inferences, and avoiding societal biases (e.g., The models are biased towards using male pronouns when presented with descriptions of careers or scientific fields). These limitations are sources of either over-generalizations or under-generalizations of learned text patterns.

Wu et al.~\cite{wu2023survey}, examine recent advancements in LLM-generated text detection, emphasizing the critical need for further research in this area. Their study analyzes commonly used datasets for LLM-generated text detection, identifying their limitations and emphasizing the need for further development. It explores key challenges faced by current detectors, such as data ambiguity and potential attacks, while also suggesting promising directions for future research to improve detection methods.

Dong et al.~\cite{dong2022survey}, provide a comprehensive overview of In-Context Learning (ICL). It presents a clear definition of ICL and its relationship to other research areas. Their survey also explores advanced techniques like training strategies, demonstration design methods, and relevant analyses used in ICL research. Addressing the limitations of current research it also suggests promising areas for future exploration.

This survey~\cite{fields2024survey} explores the diverse applications and considerations for using powerful LLM models in text classification tasks. It breaks down different types of text classification and analyzes the performance of LLMs on various datasets. While it presents its impressive capabilities, the study emphasizes the need for careful evaluation and consideration of cost, access, and ethical implications when deploying LLMs in real-world settings.

\begin{figure}[h]
  \centering
  \includegraphics[width=\linewidth]{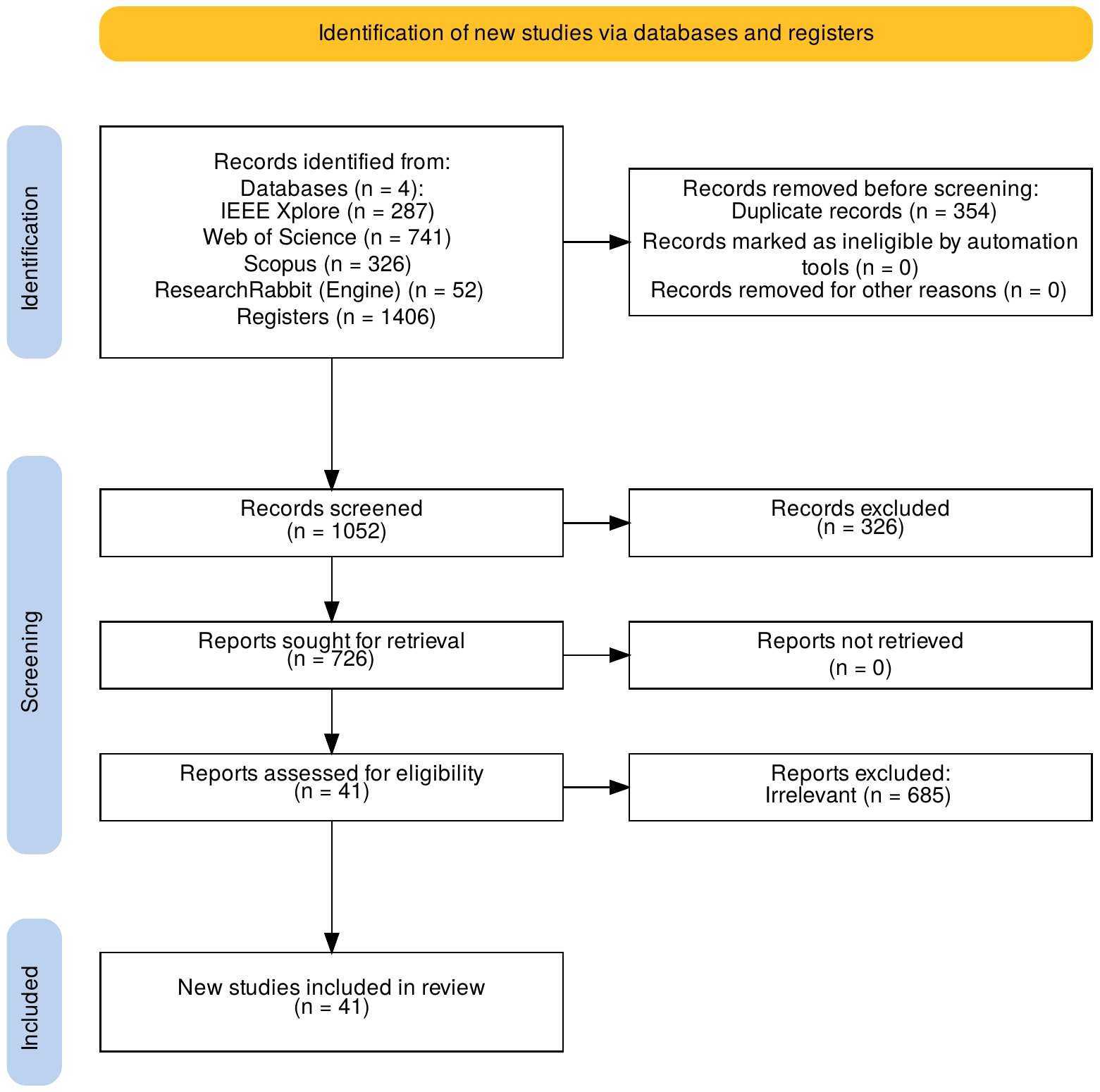}
  \caption{The PRISMA 2020 flow diagram of the performed SLR.}
  \Description[Number of retrieved, included, and excluded studies.]{This figure compares the number of retrieved studies across databases/search engines included and excluded in the SLR.}
  \label{fig2}
\end{figure}

In this SLR, we investigate studies published between 2018 and January 2024. To the best of our knowledge, this is the first SLR to analyze research on PLMs for domain-specific applications. Following the PRISMA statement (presented in Fig.~\ref{fig2}), we reviewed 41 articles. The review begins with the evolution of text classification methods, followed by an overview of both modern and traditional approaches. Subsequently, the review delves into the unique challenges and considerations inherent to domain-specific text mining. The study further categorizes existing research based on various PLMs and proposes a taxonomy of techniques used in the field. Furthermore, we conduct experiments to present and visualize the performance of general-purpose (BERT) and domain-specific (SciBERT, BioBERT) PLMs in the classification of biomedical sentences. It also addresses potential future research directions and acknowledges the limitations of this study.

\subsection{Research Methodology}
\label{subsec:research_methodology}
In this SLR, we adhere to the PRISMA statement, a well-recognized framework for conducting and reporting systematic reviews. This SLR highlights the significant progress made in text classification, with a particular emphasis on advancements in domain-specific contexts (presented in Fig.~\ref{fig1}).

\emph{Eligibility criteria:} This review offers a comprehensive analysis of recent advancements in leveraging PLMs for domain-specific text classification and techniques utilized in the field. To maintain a targeted approach, the review adheres to strict inclusion criteria. Only studies published between 2018 and January 2024 are considered, excluding studies that do not explicitly mention either ``text classification,'' ``language models,'' ``pre-trained language model,'' or ``large language models'' in their titles, abstracts, or keywords.

\emph{Information sources:} 
A diverse range of resources was utilized, including established academic databases like IEEE Xplore, Web of Science, and Scopus, alongside specialized reference management tools like Zotero. Additionally, online platforms designed specifically for literature exploration and selection empowered by AI, such as Rayyan and researchrabbit.ai, were utilized to efficiently identify and manage the studies. The last search was conducted on March 15\textsuperscript{th}, 2024.

\emph{Search strategy:}
To streamline data collection and study selection, we utilized two web-based AI tools: ResearchRabbit~\cite{researchrabbit} and Rayyan~\cite{rayyan}. Across all databases and websites, we prioritized studies specifically designed and trained on domain-specific corpora. We employed tailored queries in each source (details in Table~\ref{tab:queries}) to extract relevant results, which were then fed into Rayyan. This platform's AI capabilities efficiently assisted in choosing desired studies and excluding irrelevant ones.

\begin{table}
  \caption{Research Queries Executed}
  \label{tab:queries}
  \begin{tabular}{ll}
    \toprule
No. & Research Query\\
\hline
RQ 1 & LLM Text Classification\\
\hline
RQ 2 & LLM Scientific Text Classification\\
\hline
RQ 3 & LLM Domain-Specific Text Classification\\
\hline
RQ 4 & Large Language Model Scientific Text Classification\\
\hline
RQ 5 & Large Language Model Domain Specific Text Classification\\

\bottomrule
\end{tabular}
\end{table}

\emph{Selection process:} To ensure accuracy in our SLR, we employed strict inclusion criteria focused on PLMs and techniques used in domain-specific text classification. We first conducted an initial selection based on these criteria. Subsequently, all studies underwent Rayyan's ``compute rating'' function, followed by a final review by the authors to double-check the inclusion and exclusion of studies.

\emph{Data Items:}
\begin{itemize}
    \item Primary Outcomes: We investigated the performance of PLMs in various domain-specific text classification tasks. Our primary focus was on metrics like accuracy, precision, recall, and F1 score. We aimed to collect all compatible results for these outcome domains within each study, encompassing different measures, time points, and analyses employed by the study authors. This allowed us to comprehensively assess the effectiveness of PLMs across diverse tasks, including Named Entity Recognition (NER), RE, and Question Answering (QuA).
    \item Secondary Outcomes: These included the effectiveness of different PLMs in handling domain-specific text, the impact of specialized vocabulary and grammatical structures on model performance, and the challenges related to imbalanced data distributions.
\end{itemize}

\emph{Data collection process:}
To efficiently identify relevant studies, we utilized ResearchRabbit, an AI-powered web tool that assists researchers in finding publications across various digital libraries and allows exporting results to reference management tools like Zotero. ResearchRabbit leverages SemanticScholar for its search and prioritizes focus, displaying only the top 50 results per query. We initially employed our search queries on ResearchRabbit and added the most relevant findings to our collection. Subsequently, the same queries were applied to digital libraries like IEEE Xplore, Web of Science, and Scopus (details in Table~\ref{tab:retrieved_studies}). For each selected study, a precise review was conducted, exploring the details of the employed methodologies, followed by the creation of concise summaries for each. Finally, key data directly addressing the research questions were extracted from the summaries.

\emph{Study Risk of Bias Assessment:}
\begin{itemize}
    \item \textit{Automation Tools:}  Automation tools played a crucial role in streamlining the risk of bias assessment process. Specifically, we utilized Rayyan and ResearchRabbit, AI-powered tools designed for systematic reviews, to facilitate the initial screening and rating of studies. The functionalities of these tools included:

    \begin{itemize}
        \item AI-Assisted Screening: Both Rayyan and ResearchRabbit's AI algorithms helped in quickly screening large volumes of studies, prioritizing those that met the inclusion criteria based on keywords and abstracts.
        \item Duplicate Detection: Rayyan efficiently identified and removed duplicate records, ensuring a cleaner dataset for review.
        \item Conflict Resolution: Rayyan, in particular, flagged studies with conflicting inclusion/exclusion decisions by different reviewers, highlighting them for further discussion and resolution.
    \end{itemize}

    \item \textit{Reviewing Process:} To ensure a thorough and unbiased assessment, a multi-step reviewing process was implemented:
    \begin{itemize}
        \item Independent Review: Two reviewers independently assessed each study to determine the risk of bias. This independent evaluation helped minimize individual reviewer bias and provided a more objective assessment.
    \end{itemize}
\end{itemize}

\emph{Independent Review and Consensus:} During the initial assessment phase, each reviewer conducted their evaluations independently to ensure an unbiased analysis. Upon completing their assessments, the reviewers convened to compare and discuss their findings. In instances where discrepancies or disagreements arose, the reviewers engaged in thorough discussions to reach a consensus.

\emph{Synthesis Methods:}
\begin{itemize}
    \item Characteristics and Risk of Bias: This SLR analyzed 41 articles from 2018 to January 2024, covering diverse fields such as biomedical, radiology, computer science, finance, nuclear, material, and social sciences. The primary focus was on the application of PLMs for domain-specific text classification, encompassing techniques like traditional machine learning, rule-based methods, and modern approaches such as transformer-based models, zero-shot, few-shot learning, and Chain-of-Thought prompting. Despite using strict inclusion criteria and AI tools (ResearchRabbit and Rayyan) for screening, there remains some risk of bias due to potential publication bias and reliance on AI tools.
    \item Investigation of Heterogeneity: Significant performance variability was observed across domains, influenced by factors like specialized vocabulary, unique grammatical structures, and imbalanced data distributions. Techniques like transfer learning and prompt-based learning, including zero-shot and few-shot learning, showed varying degrees of effectiveness depending on task complexity and subjectivity. High computational costs, ethical considerations, and domain adaptation challenges contributed to the heterogeneity in model performance.
    \item Sensitivity Analyses: Sensitivity analyses focused on the robustness of PLMs across different domains by examining the impact of model size, training data volume, and fine-tuning strategies. Activation fine-tuning was particularly effective for legal text classification tasks, showing robustness against variations in domain-specific data. The effectiveness of different prompt engineering strategies was also analyzed, highlighting the importance of task-specific prompt design for optimizing model performance.
    \item Assessments of Certainty: The review provides high certainty in the effectiveness of PLMs for domain-specific text classification, supported by consistent improvements in accuracy, precision, recall, and F1 scores across multiple studies and domains. The rigorous methodology, including AI-assisted screening and detailed comparative analyses, strengthens the evidence. However, certainty is moderated by performance variability across different domains and challenges with domain-specific adaptations. The review underscores the need for ongoing research to address ethical considerations, computational costs, and refine techniques for optimizing PLM performance in domain-specific contexts.
    
\end{itemize}

\begin{table}
  \caption{Studies Retrieved per Database / Search Engine}
  \label{tab:retrieved_studies}
  \begin{tabular}{ll}
    \toprule
Database / Search Engine & Total\\
\hline
IEEE Xplore & 287\\
Web of Science & 741\\
Scopus & 326\\
ResearchRabbit & 52\\
\hline
Total & 1406\\

\bottomrule
\end{tabular}
\end{table}

\section{Evolution of Text Classification Methods}
\label{sec:evolution_of_text_classification_methods}
\begin{figure}[h]
  \centering
  \includegraphics[width=\linewidth]{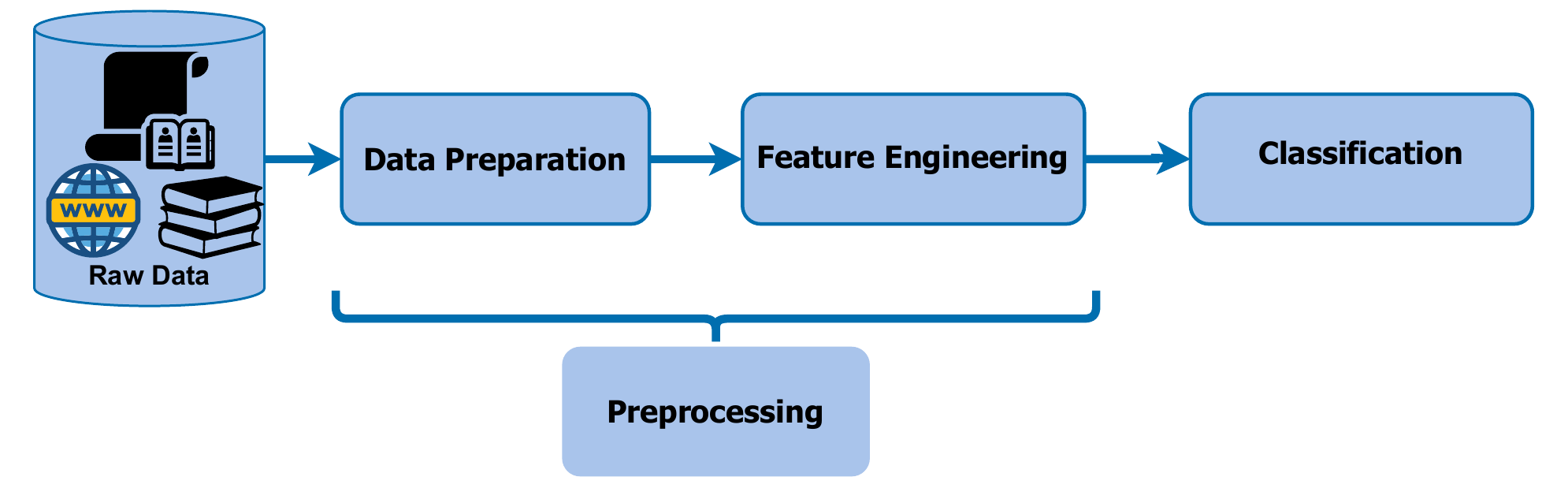}
  \caption{Text classification basic steps.}
  \Description[Main stages of text classification.]{This figure presents core steps involved in text classification process.}
  \label{fig3}
\end{figure}

The domain of text classification has experienced a notable transformation throughout its evolution~\cite{luo2023exploring, garrido2023comparing}, shifting from initial reliance on rule-based methods and regular expressions to traditional machine learning methods, and ultimately advancing to deep neural networks. This progression reflects a significant leap in the capabilities and applications of NLP. The text classification process involves four main steps: data collection, data preparation (cleanup), feature engineering~\cite{parizi2023comparative}, and label classification (presented in Fig.~\ref{fig3}). These steps collectively contribute to the effectiveness and accuracy of text classification systems. Data collection refers to gathering a large corpus of text data relevant to the classification task. The data preparation step encompasses a set of processes for cleaning up the data and preparing for the next step. Tasks have been taken care of at this level such as: noise and stop word removal, capitalization, tokenization, etc. In the feature engineering step text data should be converted to a form understandable by machine. Recently, word embeddings have been a commonly used technique to transform text to discrete text vectors using pre-defined vocabulary~\cite{beltagy2019scibert, levine2019sensebert, turian2010word, devlin2018bert}. Based on the extracted features (hand-crafted) various types of machine learning or DL classifiers can be used~\cite{abriefsurvery, parizi2023comparative}. 

There are three main groups of classifiers for text classification. First, before the advent of deep learning, researchers for class prediction utilized shallow classification methods (Naïve Bayes, k-Nearest Neighbors (kNN), Support Vector Machine (SVM), Random Forest, and Conditional Random Field), these types of classifiers have limitation in effectiveness, lack of contextual information and word dependencies~\cite{ahanger2022novel, abriefsurvery, morales2022comparison}. Second, when deep learning techniques developed and became popular, many NN-based methods were utilized in text classification. Compared to the previous approach, these models achieved better results and can learn effective text representation and prediction. Among them, models using Recurrent Neural networks (RNNs) represented by Long Short Term Memory (LSTM)~\cite{prabakaran2023bidirectional}, and Gated Recurrent Unit (GRU) have achieved notable progress in text classification. However, these models are often less efficient when handling long text documents and demand more data~\cite{ahanger2022novel, abriefsurvery, zhao2021self}. Recently, deep learning technologies have become more and more advanced. Transformers can be classified as a third group, as a powerful deep learning architecture that has been widely used~\cite{devlin2018bert, sun2023text, wang2023text2topic, ma2022pretrained, luo2022biogpt, lee2020biobert, shen2023sscibert, beltagy2019scibert, yuan2022biobart, kanakarajan2021bioelectra}. Transformer-based models have achieved significant success and recognition in various fields including text classification~\cite{beltagy2019scibert, devlin2018bert, prabakaran2023bidirectional, sun2023text, wang2023text2topic, ma2022pretrained}.

\section{Modern and Traditional Text Classification Approaches}
\label{sec:modern_and_traditional_text_classification_approach}
In this section, we will focus on various approaches utilized for text classification, particularly domain-specific text classification. We mainly emphasize transformer-based models.

\subsection{Traditional Approaches for Text Classification}
\label{subsec:traditioan_approaches_for_text_classification}
Numerous studies have explored text classification techniques. This section presents comparative analyses of state-of-the-art (SOTA) methods utilizing traditional machine learning approaches as a classifier for domain-specific and general-purpose text classification. In recent years there has been an exponential growth in the volume of available Internet data~\cite{beltagy2019scibert, wang2023text2topic}. Various initiatives are currently underway to develop tools capable of efficiently discovering, filtering, and managing these electronic resources, with an emphasis on accuracy and performance. Text classification has a direct impact on the advancement of automated tools and has a major role in their applications across diverse domains~\cite{abriefsurvery, garrido2023comparing, soni2023comparing}. 

In the context of IT service management, for example, text classification is employed for the purpose of categorizing IT support tickets. These tickets represent issues encountered by customers and are submitted as bug reports to the support team. The manual categorization of these incoming bug reports previously demanded a substantial amount of agent time. Recently, LLMs have achieved SOTA results in numerous NLP tasks, including text classification. These attention-based models have a large number of trainable parameters, allowing them to capture and represent complex linguistic patterns. Despite the computational demands associated with LLMs, there is a growing trend toward employing these models such as PLMs for NLP tasks.

Wahba et al.~\cite{wahba2023attention}, underscored the importance of carefully assessing the applicability of such models in industrial settings, particularly those that are domain-specific. The study acknowledges the effectiveness of DL models for classification tasks and argues that within a domain-specific context, many technical terms carry more precise meanings. A comparison is provided between some of the reported SOTA results in the literature and a LinearSVM classifier combined with Term Frequency-Inverse Document Frequency (TF-IDF) vectorization. The datasets used for comparison include 20News Group, BBC News, and IT support tickets. The results indicate an accuracy of 90.0\%, 98.0\%, and 79.0\% when employing SVM+TFIDF for each dataset, respectively. The study concludes that for domain-specific text classification tasks, linear models can serve as a comparable, cost-effective, reproducible, and interpretable alternative to attention-based models. 

Garrido-Merchan et al.~\cite{garrido2023comparing}, presented a comparison study between traditional ML NLP models with the most popular PLM (BERT) model for classifying text. The study investigated four various experiments in text classification including: IMDB movie reviews, RealOrNot tweets, Portuguese news, and Chinese hotel reviews. Two different classifiers were examined in each. Building bag-of-word is one of the classical supervised learning methods that is employed within NLP tasks, it uses the TF-IDF algorithm to identify the most important words in the text, this method has been used as a classical approach in the study. In contrast, BERT has demonstrated SOTA performance on a wide range of NLP tasks. This model is composed of two main steps (see Fig.~\ref{fig4}), pre-training and then fine-tuning. Unlabeled data is used for training the model on the large corpora in the phase of pre-training. Then, in the fine-tuning phase, the model was initialized with parameters that were received from the pre-training phase and fine-tuned the model with labeled data for specific tasks based on the received parameters. The study shows that while utilizing TF-IDF to extract features and train a machine learning model, BERT could outperform traditional NLP approaches and achieved these accuracies: 0.9387, 0.83640, 0.91196, and 0.9381 for IMDB movie reviews, RealOrNot tweet, Portuguese news, and Chinese hotel reviews tasks, respectively. 

\subsection{Modern Approaches for Domain-Specific Text Classification}
\label{subsec:modern_approaches_for_domain_specific_text_classification}
\begin{figure}[h]
  \centering
  \includegraphics[width=\linewidth]{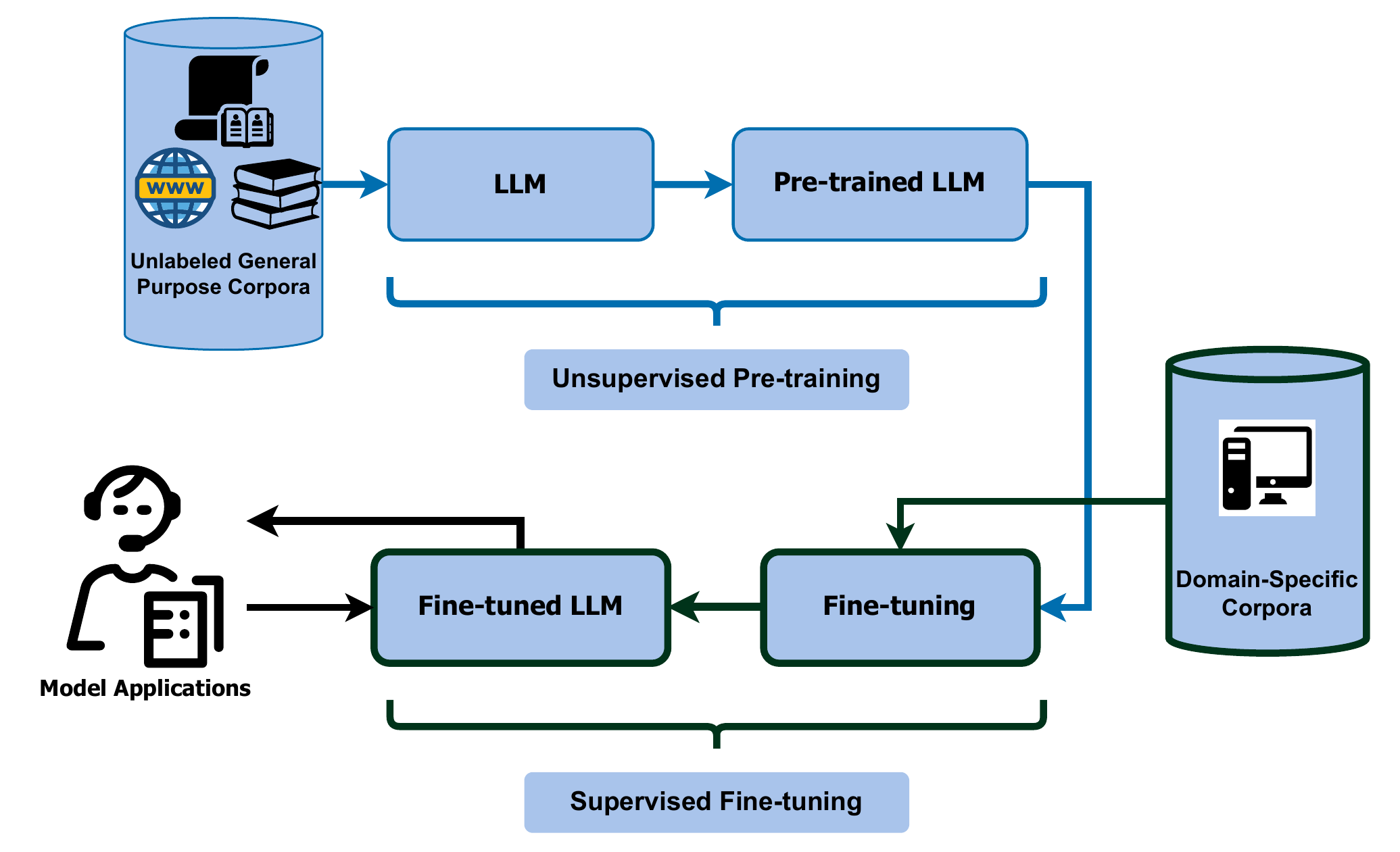}
  \caption{Pre-train and Fine-tune with LLM.}
  \Description[Utilizing PLMs for domain-specific text classification.]{PLMs can be further fine-tuned for specific applications, such as generating different text formats.}
  \label{fig4}
\end{figure}

Recently, LLMs which are essentially Transformer-based language models with hundreds of billions or more parameters become a popular tool for most NLP tasks~\cite{jeong2022scideberta, kim2023medibiodeberta, lehman2023we, zhao2023survey}. Transformer is a deep learning model designed based on an attention network, it has achieved a remarkable performance in various NLP benchmarks~\cite{devlin2018bert, sun2023text, zhai2023bytetransformer}. However, LLM performance in text classification is still significantly low compared to fine-tuned models, due to the inability to deal with reasoning and understand complex linguistic phenomena, and the restriction on token length in ICL. Examples of LLMs include GPT-3, PaLM, Galactica, and LLaMA~\cite{sun2023text, zhao2023survey}. Based on the model architecture LLMs can be broadly classified into three main categories: First, encoder-only models, which use masked language models to uniformly mask some percentage of tokens randomly and then try to predict [MASK] tokens~\cite{devlin2018bert}. They follow the pre-training and then fine-tuning paradigm for NLP tasks. Second, decoder-only models, which use the decoder of an autoregressive transformer for predicting the next token in the sequence based on preceding tokens in the sequence, also follow the pre-training then fine-tuning paradigm. Third, encoder-decoder transformer models generate new text based on given input and following the pre-training then fine-tuning paradigm~\cite{vaswani2017attention} (Pre-train and Fine-tune presented in Fig.~\ref{fig4}).

\subsection{Domain-Specific Text Classification Challenges and Consideration}
\label{subsec:domain_specific_text_classification_challenges_and_consideration}
While LLMs have demonstrated remarkable capabilities in text classification, their deployment comes with difficulties and requires careful consideration of various factors. In this section, we will briefly address challenges and considerations for domain-specific NLP tasks.

\subsubsection{Computational Resources and Inference Speed}
\label{subsubsec:computational_resources_and_inference_speed}

One of the primary challenges in deploying LLMs is the substantial computational resources required for both training and inference. Training these models demands powerful hardware and significant time and deploying them for real-time inference can be computationally intensive~\cite{yang2023training, zhai2023bytetransformer, sheng2023flexgen, huang2019gpipe, aminabadi2022deepspeed, spector2023accelerating}. 

\subsubsection{Ethical and Bias Concerns}
\label{subsubsec:ethical_and_bias_concerns}
The ethical considerations surrounding the utilization of LLMs in text classification systems have attracted considerable attention. LLMs trained on extensive and varied datasets may unintentionally reinforce biases present in the training data. Addressing bias and guaranteeing fairness in classification results outcomes are crucial~\cite{chang2024language, abid2021persistent, nangia2020crows}.

\subsubsection{Domain adaptation and Transfer Learning}
\label{subsubsec:domain_adptation_and_transfer_learing}
Text classification tasks often involve adapting models to specific domains or fine-tuning domain-specific data. While LLMs offer pre-trained representations that capture general linguistic patterns, adapting them to domain-specific nuances requires careful consideration~\cite{usha2022named}.

\begin{table}
  \caption{Summary on Reviewed PLMs for Domain-Specific Tasks.}
  \label{tab:summery_retrieved_studies}
  \scalebox{0.7}{
  \begin{tabular}{p{70pt}p{240pt}p{270pt}}
    \toprule

Studies&
Aims&
Outcomes \\
\hline
BioBERT~\cite{lee2020biobert} & This study evaluates BioBERT, a biomedical LM, on \textbf{various biomedical text mining tasks} (NER, RE, QuA) against SOTA models.  & 
BioBERT outperforms other models in biomedical text mining tasks (NER, RE, QuA). Specifically, it improves the micro-average F1 score by 0.62, 2.80, and 12.24 for NER, RE, and QuA tasks respectively.
In the RE task, the study employs the sentence classifier from the original BERT model, which utilizes the [CLS] token for classification purposes.
\\
\hline

MediBioDeBERTa~\cite{kim2023medibiodeberta} & This study proposes two novel \textbf{biomedical} models SciDeBERTa v2 and MediBioDeBERTa-IFT. Enhance the performance of the existing SOTA models in NER, semantic similarity, QuA, and RE. & SciDeBERTa v2 achieves SOTA performance in NER tasks. MediBioDeBERTa-IFT improves on existing models in NER, semantic similarity, and QuA by 0.28-1.29\% (Micro-avg-F1). It also achieves a 14.96\% improvement (F1 score: 93.04) on the ChemProt RE task.
\\
\hline

BioALBERT~\cite{naseem2022benchmarking} & A new language model tool for \textbf{biomedical and clinical NLP tasks}. They evaluate its effectiveness against current BioNLP models on six different tasks using various benchmark datasets.  & 
The model achieves SOTA performance on five out of six tasks including NER, RE, semantic similarity, document classification, and QuA. It achieves Micro avg-F1: 87.92 on HoC dataset for document classification.
\\
\hline

BioELECTRA~\cite{kanakarajan2021bioelectra} & Introduce a powerful tool for \textbf{biomedical text}. The model evaluate effectiveness across multiple tasks relevant to biomedical text analysis. & 
The model outperforms existing methods on seven key tasks: NER, PICO, RE, semantic similarity, document classification (multi-label, achieving 83.50 Micro-F1), QuA, and NLI.
\\
\hline

BioBART~\cite{yuan2022biobart} & Leveraging BART for the \textbf{biomedical domain}, encompass various biomedical tasks such as dialogue, summarization, entity linking, and NER. & 
Examines how sentence permutation affects downstream tasks and showcases BioBART's strong performance. It achieves impressive BERTscore metrics of 0.85 and 0.933 for dialogue and summarization tasks, with recall rates of 93.26\% (rank 1) and 95.74\% (rank 2) on entity linking tasks using the BC5CDR benchmark.
\\
\hline

RadBERT~\cite{yan2022radbert} &  RadBERT specifically designed for \textbf{radiology tasks}. It is trained in two steps. The study assess the performance of RadBERT variants in three key NLP tasks in radiology: \textbf{abnormal sentence classification}, report coding, and report summarization.
& All RadBERT variants demonstrate high performance in abnormal sentence classification, achieving accuracies above 97.5\% and F1 scores exceeding 95.0\%.
\\
\hline

SciBERT~\cite{beltagy2019scibert} & A new PLMs trained on \textbf{scientific text} to improve performance in various NLP tasks. The study evaluates SciBERT on multiple tasks, including NER, PICO Extraction, \textbf{text classification, relation classification}, and dependency parsing.
& 
Demonstrates superior performance compared to BERT-Base on both biomedical and computer science tasks. It achieves F1 score: 70.98 in the text classification task.
\\
\hline

SciDeBERTa~\cite{jeong2022scideberta} & SciDeBERTa is a new language model designed for \textbf{science and technology tasks}. It builds upon a general-purpose DeBERTa model and fine-tunes it with a massive scientific text dataset, aiming for better performance in science and technology NLP tasks. & 
Demonstrates enhanced performance compared to conventional language models such as SciBERT and S2ORC-SciBERT. SciDeBERTa (CS) trained in the computer science domain, achieve improved performance.
\\
\hline

BloombergGPT~\cite{wu2023bloomberggpt} & Develop a LLM with 50 billion parameters specifically designed for processing \textbf{financial data}. & A new financial language model shows promise in understanding financial data. Trained on financial news sentiment classification, it achieves an F1 score of 51.07, demonstrating its potential for financial NLP tasks.
\\
\hline

FinBERT~\cite{araci2019finbert} & A specialized LM designed for \textbf{financial tasks}, aiming to improve sentiment analysis within the financial domain. & 
The model achieving SOTA performance on sentiment analysis tasks. The model outperforms others on both the FiQA and Financial PhraseBank datasets, reaching an F1 score of 0.95 on the latter. This suggests strong capability for financial sentiment analysis.
\\
\hline

NukeBERT~\cite{jain2020nukebert} & This research builds a \textbf{nuclear-focused language model}, NukeBERT. They evaluate NukeBERT on various nuclear industry NLP tasks \textbf{(sentence classification}, NER) to see how well it performs in this specific domain. & NukeBERT significantly outperforms the base BERT model on various NLP tasks. It achieves impressive F1 and exact match scores. The study also proposes Nuke-VOCAB, a data-efficient technique, emphasizing the importance of data optimization for specialized models.
\\
\hline

NUKELM~\cite{burke2021nukelm} & NukeBERT is a LM trained on 1.5 million \textbf{nuclear research abstracts}. The researchers evaluate NukeBERT's ability to \textbf{classify research articles} into different categories within the nuclear domain. & Achieved high accuracy (0.95) and F1 score (0.81) during fine-tuning on a binary classification task.

\\
\hline

HumBERT~\cite{tamagnone2023leveraging} & Introduce a LM for \textbf{classifying entries in humanitarian data}. Importantly, the study focuses on developing methods to identify and mitigate potential biases within HumBERT and similar models trained on humanitarian text data. & 
The model outperforms generic models, especially in tasks with multiple categories.  It achieves an average F1 score of 0.73, demonstrating its effectiveness for humanitarian data analysis.
\\
\hline

SsciBERT~\cite{shen2023sscibert} & The study introduces SsciBERT, a PLM tailored for \textbf{social science text}, aiming to capture its unique characteristics. It evaluates SsciBERT's performance in \textbf{classification}, identification, and recognition tasks, comparing it with baseline models like BERT-base and SciBERT. & 
The SSCI-SciBERT-e4 variant demonstrates the most promising performance among all variants, achieving the highest accuracy, macro-average, and weighted average F1 score, with an average of 60.75
\\
\hline

MatSciBERT~\cite{gupta2022matscibert} & A specialized language model trained on a vast dataset of scientific literature within the \textbf{materials science domain}. Assess MatSciBERT's effectiveness on three downstream tasks: \textbf{abstract classification}, named entity recognition (NER), and relation extraction (RE), across diverse materials datasets.& 
MatSciBERT outperforms SciBERT, a general-purpose language model, across all evaluated tasks. Notably, it achieves a high F1 score of 93.57\% in abstract classification, demonstrating its efficacy in distinguishing manuscripts based on glass versus non-glass topics.\\

\bottomrule

\multicolumn{3}{p{600pt}}{The use of bold text within the ``Aims'' column highlights the model's core area of expertise or the specific tasks it was created to handle. }

\end{tabular}
}
\end{table}

\subsection{Pre-trained Language Models for Domain-Specific Text Classification}
\label{subsec:pre_trained_language_models_for_domain_specific_text_classification}
PLMs are adaptable tools capable of handling a wide range of tasks within NLP. These tasks include text classification, text summarizing, question answering, text generation, sentiment analysis, NER, and textual entailment. PLMs are trained on extensive text data, excel at capturing complex linguistic features and semantic relationships. However, domain-specific text mining in scientific fields faces challenges due to specialized vocabulary and complex structures. Traditional methods struggle in these domains~\cite{beltagy2019scibert, jeong2022scideberta, kim2023medibiodeberta, lee2020biobert, shen2023sscibert}. PLMs, however, provide a promising solution, leveraging pre-training knowledge to capture linguistic patterns in scientific texts, leading to improved accuracy. In this section, we explore the recent PLMs for domain-specific contexts which can be utilized for various tasks including text classification. Table~\ref{tab:summery_retrieved_studies} presents a summary on reviewed PLMs designed for domain-specific tasks.

\subsubsection{Pre-trained Language Model for Biomedical Tasks}
\label{subsubsec:pre_trained_model_for_biomedical_tasks}
In recent years, the landscape of biomedical research has experienced a remarkable advancement, largely pushed by the introduction of sophisticated NLP techniques. Among these advancements, PLMs have a major role in addressing diverse challenges within the biomedical field. In the following section, we highlight the recent influential papers that have played a significant role in advancing the application of PLMs in biomedical research (see Table~\ref{tab:summery_biomedical} for biomedical PLMs summaries, and Fig.~\ref{fig5} for biomedical PLMs comparison on NER tasks).

Lee et al.~\cite{lee2020biobert} have proposed a domain-specific pre-trained language representation (BioBERT) that trained on a massive number of biomedical corpora.  BioBERT achieved better results in a variety of text mining tasks and outperformed SOTA models. The model has been designed and weights initialized based on the BERT which is trained on general domain corpora. BioBERT trained on extracted data from PubMed abstracts and PMC fulltext articles to create a model that functions within a biomedical corpus. This model is then evaluated on three representative biomedical text mining tasks: biomedical named entity recognition, biomedical relation extraction~\cite{zhao2023comprehensive}, and biomedical question answering. Notably, for the RE task, the researchers leverage the sentence classifier from the original BERT model, which utilizes the [CLS] token for classification. The study concludes that BioBERT significantly improves performance, outperforming the micro average F1 score by 0.62, 2.80, and 12.24 compared to the SOTA models for these tasks, respectively.

Kim et al.~\cite{kim2023medibiodeberta}, propose two models which are SciDeBERTa v2, and MediBioDeBERTa-IFT.  The first model, SciDeBERTa is a specialized PLM for biomedical applications. This model is trained on the S2ORC from scratch within full text and surpasses the previous version SciDeBERTa. It achieved the SOTA on the NER task in the SciERC dataset. The second proposed model is MediBioDeBERTa-IFT which improved performance over the existing SOTA models in three various tasks: NER, semantic similarity, QuA by 0.28\% (Micro-avg-F1: 86.28), 0.2\% (Micro-avg-F1: 92.7), and 1.29\% (Micro-avg-F1: 76.24) respectively, as well as the ChemProt relation extraction (RE) task on BLURB by 14.96\% (F1 score: 93.04).

Naseem et al.~\cite{naseem2022benchmarking}, introduce BioALBERT, a novel LM specifically designed for biomedical and clinical Natural Language Processing (BioNLP) tasks. BioALBERT is based on the ALBERT architecture but is optimized and adapted for the biomedical domain by training on datasets from biomedical and clinical text. Researchers investigate the effectiveness of BioALBERT by comparing its performance to existing BioNLP models on six different tasks using a diverse set of benchmark datasets. A large variant of BioALBERT trained on PubMed data achieves SOTA performance on five out of six tasks, as measured by the BLURB score. This demonstrates significant improvements in different BioNLP areas, including NER, RE, sentence similarity, document classification, and QuA. BioALBERT outperforms previous SOTA models on 17 out of 20 benchmark datasets across various tasks. This highlights its versatility, robustness, and applicability for various BioNLP applications.

BioELECTRA~\cite{kanakarajan2021bioelectra} emerges as a powerful tool for comprehending biomedical text, outperforming current methods in various tasks. This model, built upon the foundation of ELECTRA, demonstrably outperforms existing techniques in understanding medical information. Trained on a comprehensive dataset of articles from PubMed and PMC databases, BioELECTRA achieves remarkable results across seven critical tasks (NER, PICO, RE, sentence similarity, document classification, QuA, and NLI). In the document classification task, the model assigns multiple relevant labels to a text document. Additionally, it surpasses previous models on established benchmarks and achieving a new SOTA accuracy.

Yuan et al.~\cite{yuan2022biobart} introduce BioBART, a generative language model tailored for the biomedical domain by leveraging BART. BioBART encompasses various biomedical tasks like dialogue, summarization, entity linking, and NER. The study evaluates the model by utilizing different datasets and metrics like BLEU~\cite{papineni2002bleu}, ROUGE~\cite{lin2004rouge}, and BERTscore~\cite{zhang2019bertscore}, the model demonstrates SOTA results across several NLG tasks. Utilizing text-infilling, BioBART adapts BART to the biomedical domain through continuous pre-training on PubMed abstracts. The study reveals that while sentence permutation noise improves summarization tasks, it leads to slight counter impact in other biomedical NLG tasks, prompting the use of only text infilling for pre-training BioBART. Highlighting superior performance, BioBART based on BERTscore metrics achieves 0.85 and 0.933 for dialogue and summarization tasks, respectively. In entity linking tasks, BioBART attains a recall of 93.26\% (rank 1) and 95.74\% (rank 2) on the BC5CDR benchmark.

\begin{table}
  \caption{Biomedical PLMs Details for Text Classification.}
  \label{tab:summery_biomedical}
  \scalebox{0.85}{
  \begin{tabular}{p{120pt}llp{110pt}l}
    \toprule
Application & Works & Tasks \hspace{1.5cm} Result & Data Source\\ \hline
Biomedical Text Mining &  BioBERT~\cite{lee2020biobert} &

\begin{tabular}{p{55pt}p{75pt}}
     NER &  In Drug/chem type. Dataset: BC5CDR. F1 score: 93.47.\\
     \hline
     RE & In Gene-disease relation. Dataset: GAD. F1 score: 79.83.\\
     \hline
    QuA  & Dataset: BioASQ5 Lenient Accuracy: 60.0 
\end{tabular}

& PubMed abstract, and
PMC full text\\

\hline

Biomedical language understanding and reasoning benchmark (BLURB), Specialized in Scientific Domain &  MediBioDeBERTa~\cite{kim2023medibiodeberta} &

\begin{tabular}{p{55pt}p{75pt}}
     NER &  Micro-avg-F1: 86.28\\
     \hline
     Semantic Similarity & Micro-avg-F1: 92.7\\
     \hline
     QuA &  Micro-avg-F1: 76.24\\
     \hline
    RE & F1 score: 93.04\\
\end{tabular}

& S2ORC and
SciERC
\\
\hline

A novel LM specifically designed for biomedical and clinical BioNLP  &  BioALBERT~\cite{naseem2022benchmarking} &

\begin{tabular}{p{55pt}p{75pt}}
     NER &  Dataset: BLURB. Micro-avg-F1: 95.70\\
     \hline
     RE &  Dataset: BLURB. Micro-avg-F1: 79.94\\
     \hline
     Sentence Similarity &  Dataset: BLURB. Pearson-avg-score: 89.25\\
     \hline
      Document Classification & Dataset: HoC. Micro-avg-F1: 87.92\\
     \hline
      QuA & Dataset: BLURB. Lenient-score: 58.03\\
\end{tabular}

& PubMed abstract, PMC full text, and clinical notes (MIMIC) and their combination
\\
\hline

Biomedical domain-specific encoder model  &  BioELECTRA~\cite{kanakarajan2021bioelectra} &

\begin{tabular}{p{55pt}p{75pt}}
     NER &  Dataset: BC5-chem. F1 score: 93.75\\
     \hline
     PICO &  Dataset: EBM PICO. Macro-F1: 74.26\\
     \hline
     RE &  Dataset: GAD. Micro-F1: 85.67\\
     \hline
      Sentence Similarity & Dataset: BIOSSES. Pearson-avg-score: 93.80\\
     \hline
      Document Classification &  Dataset: HoC. Micro-F1: 83.50\\
     \hline
      NLI &  Dataset: MedNLI. Accuracy: 86.34\\
     \hline
      QuA &  Dataset: BioASQ. Accuracy: 91.50\\
\end{tabular}

& PubMed abstract, PMC full text.
\\
\hline

Biomedical generative language model &  BioBART~\cite{yuan2022biobart} &

\begin{tabular}{p{55pt}p{75pt}}
     Dialogue &  Dataset: Covi19-Dialogue. BERTscore: 0.852\\
     \hline
     Summarization &  Dataset: MeQSum. BERTscore: 0.933\\
     \hline
    Entity Linking &  Dataset: BC5CDR. Recall@1/@5: 93.26/95.74\\
     \hline
      NER & Dataset: ShARe13. F1 score: 80.75\\
\end{tabular}

& PubMed abstract\\
\bottomrule
\end{tabular}
}
\end{table}

\paragraph{Pre-trained Language Model for Radiology Tasks and Evaluate Model's Performance}
\label{pragraph:utilizing_llm_for_radiology}
 Yan et al.~\cite{yan2022radbert} investigate transformer-based language models for specific applications in radiology. The study proposes a RadBERT which is created through a two-step process. First, transformer models are pre-trained on a massive dataset of radiology reports obtained from the U.S. Department of Veterans Affairs healthcare systems. Subsequently, these pre-trained models undergo fine-tuning using various initialization techniques, resulting in six distinct variants of RadBERT. These models are fine-tuned for three key NLP tasks in radiology: abnormal sentence classification, report coding, and report summarization. All models excelled in abnormal sentence classification, with accuracies above 97.5\% and F1 scores exceeding 95.0\%. RadBERT variants outperformed baselines, even with minimal training data. Additionally, all RadBERT variants outperformed baselines across all five coding systems in report coding. In report summarization, RadBERT-BioMed-RoBERTa emerged as the top performer, achieving a ROUGE-1 score of 16.18.
 
 Tan et al.~\cite{tan2023inferring}, evaluate the performance of various language models in accurately classifying cancer disease response from free-text radiology reports. Using a dataset of 10,602 computed tomography reports from cancer patients, the reports were classified into four categories: no evidence of disease, partial response, stable disease, or progressive disease. The study employs transformer models, bidirectional LSTM, CNN, and conventional machine-learning techniques for this task. The best-performing model, the GatorTron transformer, achieved an accuracy of 0.8916 on the test set and 0.8919 on the RECIST validation set. Data augmentation techniques and prompt-based fine-tuning were employed to further enhance model performance. Data augmentation improved accuracy to 0.8976, whereas prompt-based fine-tuning reduced the size of the training dataset while maintaining good performance. The study demonstrates the feasibility and effectiveness of using deep learning-based NLP models for inferring cancer disease response from radiology reports on a large scale, with transformer models showing consistent superiority.
 
 Liu et al.~\cite{liu2023exploring}, examined the performance of GPT-4, a powerful language model, on various radiology tasks involving text analysis. Compared to leading radiology-specific models, GPT-4 either outperformed or matched their performance in most cases. For tasks with clear instructions, minimal prompting (zero-shot) was sufficient for GPT-4 to significantly outperform radiology models. These tasks included comparing the similarity of sentences across time and drawing logical inferences from natural language. More complex tasks requiring an understanding of specific data formats or styles (like summarizing findings) demanded example-based prompting. With this approach, GPT-4 achieved performance comparable to the best existing models requiring extensive training data. Analysis of GPT-4's errors with a radiologist revealed a strong understanding of radiology concepts. Most errors from ambiguity in the data or incorrect labeling, with some instances requiring highly specialized knowledge. GPT-4's summaries of findings often matched the quality of human-written summaries.

\subsubsection{Pre-trained Language Model for Computer Science and Biomedical Tasks}
\label{subsubsec:pre_trained_model_for_computer_science_and _tasks}
Recently, advancement in NLP has been driven by adapting to deep neural models. However, a huge amount of labeled data is required for training such models. The collection of annotated data in scientific fields presents a significant challenge due to the high cost and the requisite expertise necessary to ensure annotation quality.

Beltagy et al.~\cite{beltagy2019scibert}, propose a PLM trained on large corpora of scientific text and designed based on BERT. The study presents an approach to enhance performance on a range of NLP tasks in the scientific domain. SCIBERT is pre-trained on the scientific text and inspired by BERT architecture. It follows the same training steps as in BERT by predicting the randomly masked tokens and checking to make sure two sentences follow each other. The study built SCIVOCAB, an advanced version of WordPiece vocabulary on the scientific corpus using SentencePiece library. The vocabulary size is 30K both cased and uncased vocabularies have been produced. The corpus size used in the study is 3.17B tokens which is made-up of 18\% papers from the domain of computer science and the rest of the corpus which is 82\% from the broad biomedical domain. The model has been investigated on various NLP tasks such as: NER, PICO Extraction (PICO), Text Classification (CLS), Relation Classification (REL), and Dependency Parsing (DEP). The model outperforms BERT-Base~\cite{devlin2018bert} on both biomedical tasks and computer science tasks, it obtained +1.92 F1 with fine-tuning and +3.59 F1 without, and +3.55 F1 with fine-tuning and +1.13 F1 without respectively. 

Jeong et al.~\cite{jeong2022scideberta}, present a specialized LM that achieved improved performance in natural language understanding (NLU) within the science and technology domain. The model is designed and initialized with the DeBERTa parameters, which is trained for general domain tasks, and further trained with science and technology corpus. Although among the top models of the SuperGLUE leaderboard, DeBERTa has the smallest model size by 78 GB, it provides the best performance compared to its model size and training data size. The S2ORC dataset was utilized to train the model which covers a wide range of scientific and technological fields. The experiments present that SciDeBERTa and SciDeBERTa(CS) (which is trained with the computer science domain) achieved improved performance compared to the conventional LMs such as SciBERT and S2ORC-SciBERT (see Table~\ref{tab:summery_science_and_technology} for summaries of biomedical and computer science PLMs, Fig.\ref{fig6} for SciBERT’s F1 scores across five tasks, and Fig.\ref{fig7} for an F1 score comparison between SciDeBERTa and SciDeBERTa (CS)).

\begin{table}
  \caption{Science and Technology PLMs Details for Text Classification.}
  \label{tab:summery_science_and_technology}
  \resizebox{\textwidth}{!}{
  \begin{tabular}{p{120pt}llp{100pt}l}
    \toprule
Application & Works & Tasks \hspace{1.5cm} Result & Data Source\\ \hline
Scientific text: Biomedical and Computer Science & 
 SciBERT~\cite{beltagy2019scibert} &

\begin{tabular}{p{55pt}p{75pt}}
     NER &  Dataset: BC5CDR
F1 score: 90.01
\\
     \hline
     PICO Extraction & Dataset: EBM-NLP
F1 score: 72.28
\\
     \hline
Text Classification  & Dataset: ACL-ARC 
F1 score: 70.98
\\
     \hline
    Relation Extraction & Dataset: SciERC
F1 score: 79.97
\\
     \hline
     Dependency Parsing & Dataset: GENIA-UAS.
F1 score: 91.99 

\end{tabular}

& SCIVOCAB
Semantic Scholar:
18\% papers
form Computer Science, 82\% papers from
Biomedical
\\
\hline

Natural Language Understanding (NLU) within the science and technology domain &  SciDeBERTa~\cite{jeong2022scideberta} &

\begin{tabular}{p{55pt}p{75pt}}
     NER &  Dataset: SciERC
SciDeBERTa 70.8 ± 0.5, SciDeBERTa (CS) 71.1 ± 0.6
\\
     & Dataset: GENIA SciDeBERTa 78.7 ± 0.3, SciDeBERTa (CS) 77.8 ± 0.3
\\
     \hline
JRE  & Dataset: SciERC SciDeBERTa 45.5 ± 0.8, SciDeBERTa (CS) 46.0 ± 0.8
\\
     \hline
    Coref & Dataset: SciERC
SciDeBERTa 56.4 ± 0.8, SciDeBERTa
(CS) 57.4 ± 0.6
\\
      & Dataset: GENIA
SciDeBERTa 47.0 ± 0.7, SciDeBERTa
(CS) 46.3 ± 0.4

\end{tabular}

& S2ORC

\\
\bottomrule
\end{tabular}
}
\end{table}

\subsubsection{Pre-trained Language Model for Financial Tasks}
\label{subsubsec:pre_trained_model_for_Financialtasks}
The finance sector produces a large volume of unorganized text information, such as news articles, financial reports, social media posts, and investor filings. Deriving valuable information from this data is essential for tasks like investment analysis, risk management, and fraud detection. However, conventional text classification techniques frequently face challenges in efficiently dealing with the specialized language, intricate sentence structures, and built-in biases found in financial texts (see Table~\ref{tab:summery_finance} for financial PLMs summaries). Authors from~\cite{wu2023bloomberggpt} presented an LLM with 50 billion parameters on a wide range of financial data. The model has been trained on a comprehensive domain-specific dataset that has 363 billion tokens sourced from Bloomberg’s extensive data sources, in addition to 345 billion tokens from the general purpose dataset, as a result, the model trained with a large corpus with over 700 billion tokens. Recently, results achieved through domain-specific models have discovered that general models cannot use an alternative instead of them. This model can be employed for a diversity of tasks, but it has demonstrated particularly strong performance in financial domains, therefore it is recommended to use by a specific model. The study aims to develop a model that is trained on both domain-specific and general-purpose data sources to achieve a better result within domain-specific tasks and provide a strong performance on general-purpose benchmarks as well.

Araci et al.~\cite{araci2019finbert}, propose FinBERT, a language model specifically designed for financial tasks. The study demonstrated FinBERT on two financial sentiment analysis datasets, achieving SOTA results on both the FiQA sentiment scoring and Financial PhraseBank datasets. Additionally, the study compares FinBERT with two other pre-trained language models, ULMFit and ELMo, for financial sentiment analysis. The study investigated various aspects of FinBERT, including the benefits of further training on financial data, preventing negative learning transfer (catastrophic forgetting), and reducing training time through selective fine-tuning, only fine-tuning a small subset of model layers to reduce training time without sacrificing performance. FinBERT achieves an impressive F1 score of 0.95 on the Financial PhraseBank dataset (see Table~\ref{tab:summery_finance}).

\begin{table}
  \caption{Finance PLMs Details for Text Classification.}
  \label{tab:summery_finance}
  \resizebox{\textwidth}{!}{
  \begin{tabular}{p{100pt}p{60pt}p{80pt}p{100pt}p{80pt}}
    \toprule
Application & Works & Tasks & Result & Data Source\\ \hline
Trained on both domain-specific and general-purpose data sources & BloombergGPT~\cite{wu2023bloomberggpt} & Support diverse set of tasks &
Remarkably outperform comparable models in both general language modeling and financial tasks. & Bloomberg’s Data sources. (FINPILE)
\\
\hline
Financial sentiment analysis & FinBERT~\cite{araci2019finbert} & Sentiment analysis &
Dataset: Financial PhraseBank F1 score 0.95 & TRC2-financial, Financial PhraseBank, and FiQA Sentiment

\\
\bottomrule
\end{tabular}
}
\end{table}

\subsubsection{Pre-trained Language Model for Nuclear Domain}
\label{subsubsec:pre_trained_model_for_nuclear_domain}
Jain et al.~\cite{jain2020nukebert}, propose NukeBERT, a specialized version of the BERT language model, designed for the nuclear domain. Due to the limited amount of publicly available data in this field, traditional NLP models often struggle in this context. NukeBERT addresses this challenge by being fine-tuned on two new datasets: NText, a collection of preprocessed nuclear research papers and theses, and NQuAD, a dataset of expert-created question-answer pairs related to nuclear topics. NukeBERT achieves impressive F1 and exact match scores of 93.87 and 88.31, respectively, representing significant improvements over the original BERT model. Additionally, the study proposes a novel data-efficient technique called NukeVOCAB to optimize the training process with less data. This success highlights the effectiveness of NukeBERT for various NLP tasks in the nuclear industry, such as sentence classification and NER. This work demonstrates the potential of adapting NLP models to specialized domains with limited data and emphasizes the importance of data optimization techniques and collaboration for advancements in the nuclear field.

Burke et al.~\cite{burke2021nukelm}, introduces NUKELM, a novel language model specifically designed for the nuclear domain. NUKELM is trained on a massive dataset of 1.5 million abstracts from the U.S. Department of Energy Office of Scientific and Technical Information (OSTI) database, allowing it to understand and process nuclear-related text effectively. The study demonstrates that NUKELM excels at classifying research articles into various categories. This capability can be used for efficient manuscript sorting, analyzing citation networks within the nuclear field, and identifying new research areas. NUKELM outperforms previous models like NukeBERT in this area by utilizing a larger and publicly available dataset. Additionally, NUKELM's architecture facilitates easy fine-tuning for diverse tasks within the nuclear domain. During fine-tuning on the binary classification task, the model achieved 0.95 and 0.81 for accuracy and F1 score respectively (see Table~\ref{tab:summery_nuclear} for nuclear PLMs summaries).

\begin{table}
  \caption{Nuclear Domain PLMs Details for Text Classification.}
  \label{tab:summery_nuclear}
  \resizebox{\textwidth}{!}{
  \begin{tabular}{p{100pt}p{60pt}p{80pt}p{100pt}p{80pt}}
    \toprule
    Application & Works & Tasks & Result & Data Source\\ \hline
PLM for nuclear domain & NukeBERT~\cite{jain2020nukebert} & NLP tasks in the nuclear industry &
F1 - score 93.87 & NText, NQuAD
\\
\hline
LM for nuclear-domain & NUKELM~\cite{burke2021nukelm} & Classifying research articles into various categories  &
F1 score 0.81 & 1.5 million abstracts from U.S. Department of Energy Office
of Scientific and Technical Information
\\
\bottomrule
\end{tabular}
}
\end{table}

\subsubsection{Pre-trained Language Model for Humanitarian Domain}
\label{subsubsec:pre_trained_model_for_humanitarian_domain}
Tamagnone et al.~\cite{tamagnone2023leveraging}, offer a comprehensive approach to practice-oriented, effective, and bias-aware solutions for entry classification in the humanitarian data. It introduces HumBERT, a specialized LLM based on BERT is trained on a diverse collection of humanitarian text sources. The study also proposes a methodical strategy for identifying and mitigating biases within these models. Experimental results demonstrate that HumBERT outperforms generic language models in both zero-shot and full-training scenarios, particularly on tasks with multiple target classes. Additionally, the study introduces the HUMSETBIAS dataset to assess the impact of specific attributes on model predictions, such as gender and country. Through targeted data augmentation techniques, biases are effectively reduced without sacrificing model performance. HumBERT achieves an average F1 task score of 0.73, highlighting its effectiveness in humanitarian data analysis.

\subsubsection{Pre-trained Language Model for Social Science Texts}
\label{subsubsec:pre_trained_model_for_social_domain}
Shen et al.~\cite{shen2023sscibert} introduce SsciBERT, a  new PLM specifically designed for social science texts. The model designed upon existing models (BERT-base and SciBERT) by further training them on a large dataset of social science abstracts collected from the Web of Science. SsciBERT outperformed the baselines in tasks like classification, identification, and recognition, demonstrating its ability to better capture the unique characteristics of social science language. The study states limitations, such as using only abstracts for training and the lack of standardized social science datasets. The SSCI-SciBERT-e4 variant achieves the highest accuracy, macro-average, and weighted average F1 score, making it the most promising model among all variants, with a weighted average of 60.75\%. SsciBERT represents a significant contribution by filling the gap for PTM in social sciences, potentially aiding future research in text mining, knowledge discovery, and understanding research trends within the field.

\subsubsection{Pre-trained Language Model for Material Domain}
\label{subsubsec:pre_trained_model_for_material_domain}
Gupta et al.~\cite{gupta2022matscibert}, introduce MatSciBERT, a specialized language model trained on a massive dataset of scientific literature from the materials science domain. MatSciBERT is evaluated on three downstream tasks: abstract classification, NER, and RE, across various materials datasets. Results show that MatSciBERT outperforms SciBERT, a general-purpose language model, on all tasks. Particularly notable is MatSciBERT's high F1 score of 93.57\% in abstract classification, distinguishing between manuscripts on glass versus non-glass topics. By leveraging knowledge from both computer science and biomedical corpora, in addition to specific materials science data, MatSciBERT demonstrates superior performance and holds promise for accelerating information extraction from materials science texts.

\subsection{Domain-Specific PLMs Key Findings and Challenges}
\label{subsec:plms_key_finingds_and_challenges}
This section delves into the key findings and challenges associated with PLMs for domain-specific text classification. These models, tailored to the unique linguistic and contextual demands of various domains such as biomedical research, computer science, finance, nuclear science, humanitarian, social science, and materials science, exhibit significant improvements in task-specific performance over general-purpose models. Despite their successes, the development and deployment of these PLMs present a range of challenges, including the need for extensive domain-specific training data, and high computational costs.

\emph{Key Findings:}
\subsubsection{Biomedical Domain}
\label{biomedical_domain_challenges_and_key_findings}
\begin{itemize}
    \item Biomedical PLMs show superior performance in biomedical tasks such as NER, RE, QuA, often significantly outperforming SOTA models.
    \item BioBERT demonstrated improvements in F1 scores for NER (+0.62), RE (+2.80), and QA (+12.24).
    \item MediBioDeBERTa-IFT achieved a 14.96\% improvement in F1 score on the ChemProt RE task.
    \item BioALBERT showed SOTA performance on five out of six BioNLP tasks and outperformed previous models on 17 out of 20 benchmark datasets.
    \item BioELECTRA achieved new SOTA accuracy in document classification and other benchmarks.
    \item BioBART excelled in dialogue, summarization, entity linking, and NER tasks using continuous pre-training on PubMed abstracts.
\end{itemize}

\begin{figure}[h]
  \centering
  \includegraphics[width=0.9\linewidth]{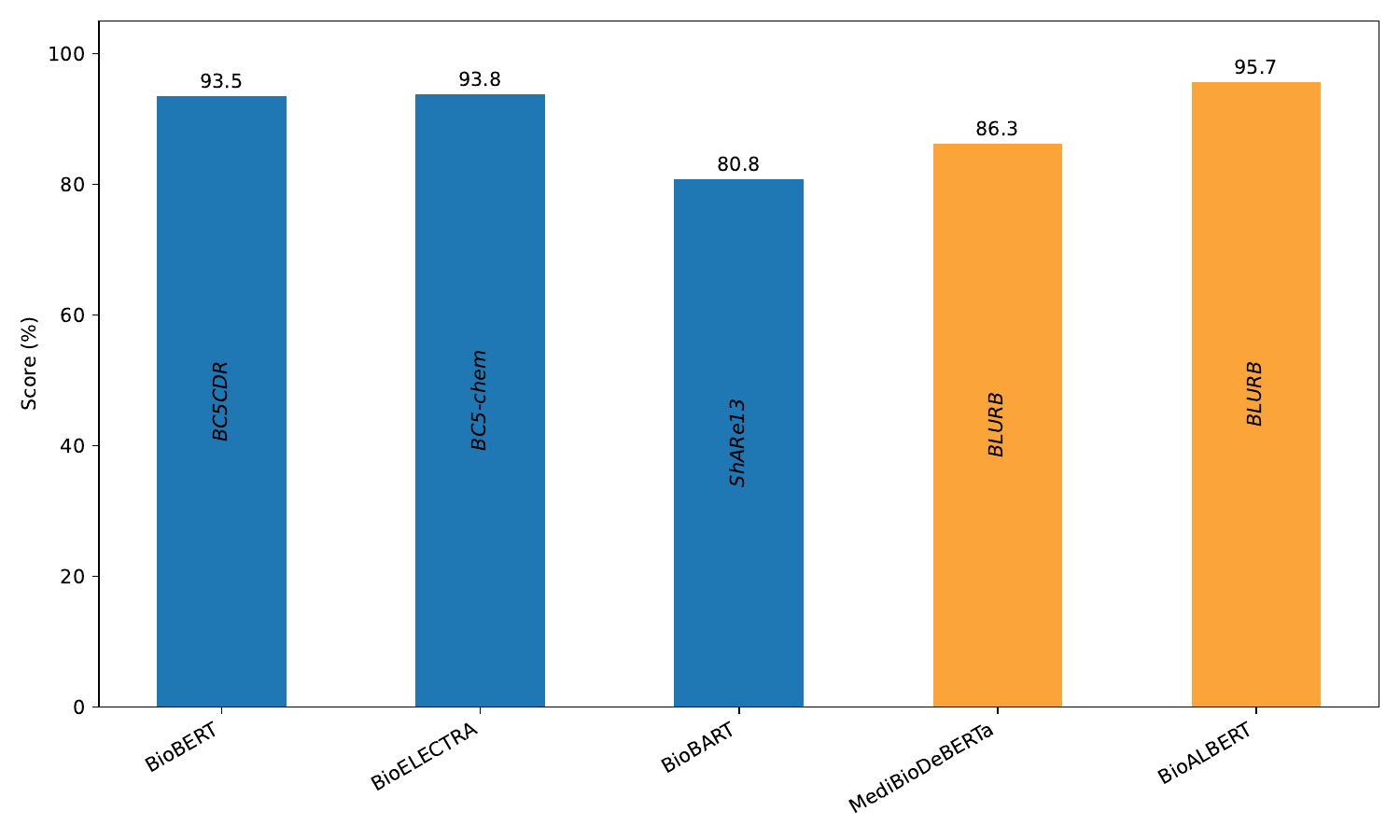}
  \caption{Comparison of biomedical PLMs on NER tasks. Models on left (blue) are evaluated using standard F1 scores on BC5CDR, BC5-chem, and ShARe13. Models on the right (orange) are evaluated using micro-averaged F1 on the BLURB benchmark.}
  \Description[Comparing different PLMs for biomedical tasks.]{The figure shows a comparison of biomedical PLMs on NER tasks, while utilizing different datasets.}
  \label{fig5}
\end{figure}

Based on the key findings and achievements highlighted by the models in Section~\ref{subsubsec:pre_trained_model_for_biomedical_tasks} and Table~\ref{tab:summery_biomedical}, it is evident that biomedical PLMs significantly outperform general-purpose models in domain-specific tasks. To illustrate the superiority of these models, Fig.~\ref{fig5} presents the performance of five biomedical PLMs (BioBERT, BioELECTRA, BioBART, MediBioDeBERTa and BioALBERT) evaluated on different NER benchmarks (BC4CDR, BC5-chem, ShARe13, and BLURB). All of the models, even though they were trained on different datasets, achieved high F-scores. The implication is that training on biomedical corpora ultimately helps in providing results. This shows that how architecture advancement and specialization from BERT to ELECTRA and ALBERT variations, contributed towards sustained performance improvements. These achievements show that domain-specific adaptation is key for achieving the best possible results in medical NLP.

\subsubsection{Computer Science and Biomedical Domain}
\label{computer_science_biomedical_domain_challenges_and_key_findings}
\begin{figure}[h]
  \centering
  \includegraphics[width=0.9\linewidth]{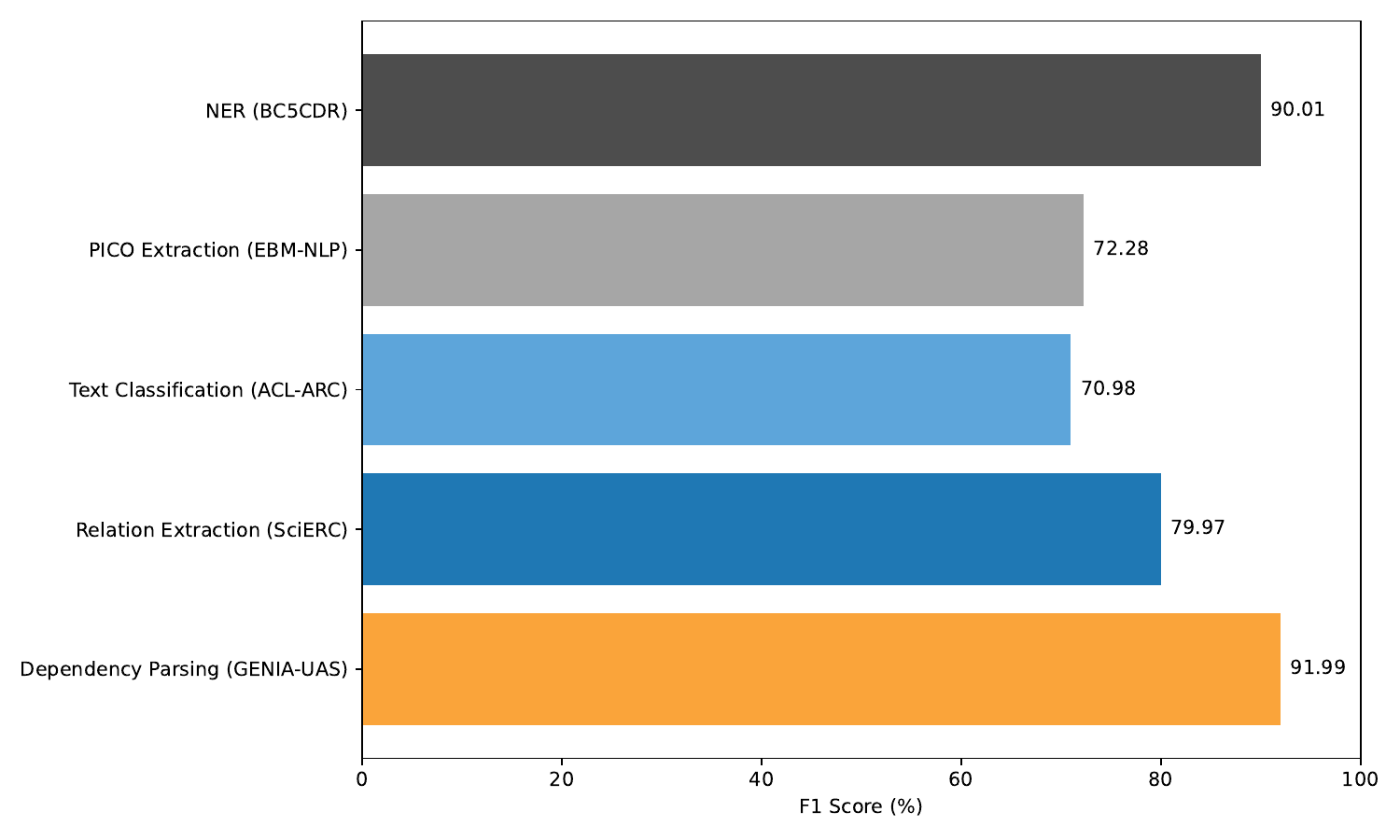}
  \caption{SciBERT F1 score on five biomedical and computer science.}
  \Description[Model performance (F1 score) across tasks.]{The figure shows a comparison of science and technology PLMs across various tasks.}
  \label{fig6}
\end{figure}

\begin{figure}[h]
  \centering
  \includegraphics[width=0.9\linewidth]{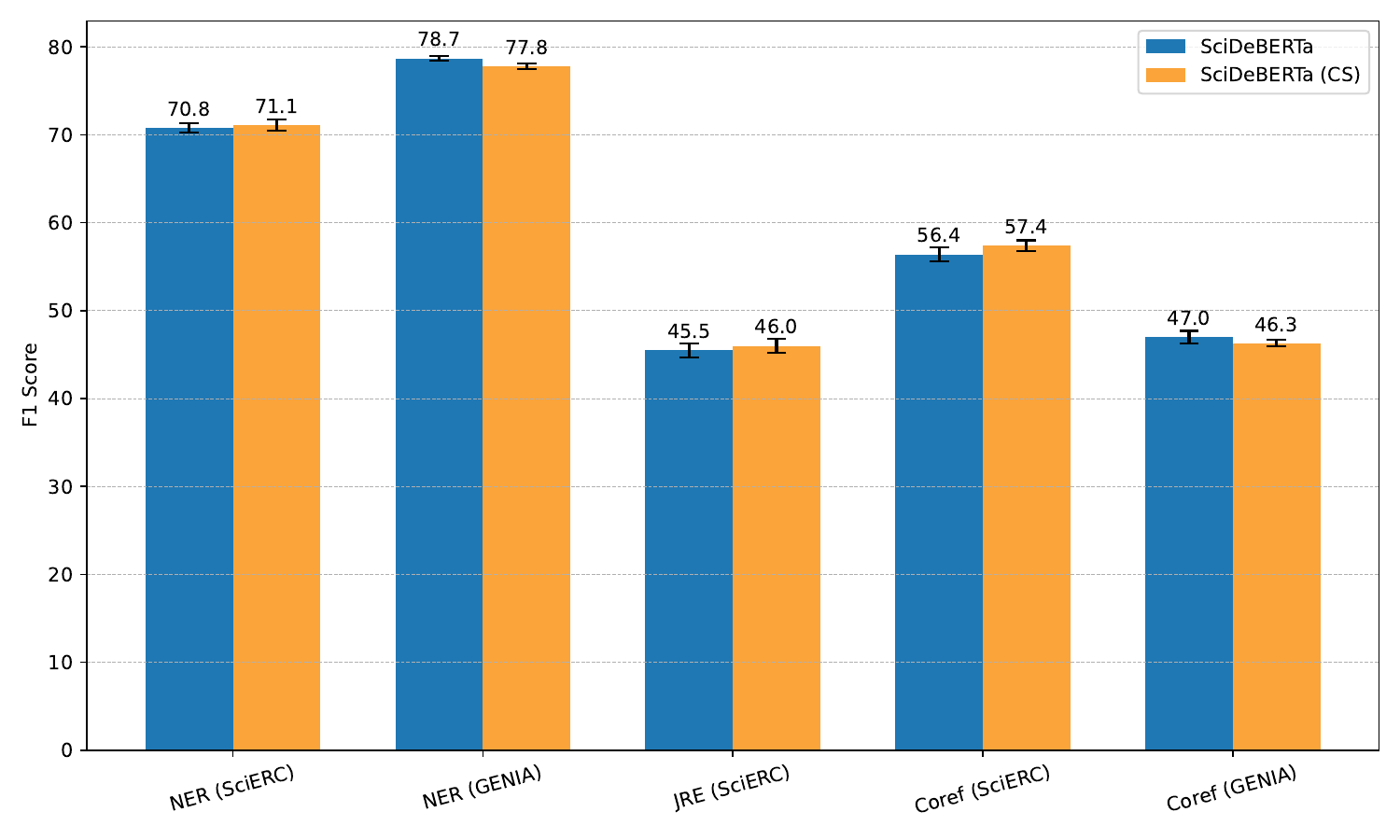}
  \caption{F1 score performance comparison between SciDeBERTa and SciDeBERTa (CS). Bars represent the mean F1 score $\pm$ standard deviation over five experimental runs.}
  \Description[Model performance (F1 score) across tasks, SciDeBERTa (CS) refers to the variant continually pre-trained on Computer Science abstracts, while SciDeBERTa is trained on the broader S2ORC corpus.]{The figure shows a comparison of between SciDeBERTa and SciDeBERTa (CS).}
  \label{fig7}
\end{figure}

\begin{itemize}
    \item SciBERT, trained on scientific texts with a specialized vocabulary (SCIVOCAB), outperformed BERT-Base in biomedical and computer science tasks (see Fig.~\ref{fig6}).
    \item SciDeBERTa, initialized with DeBERTa parameters, showed enhanced performance in NER, RE, and coreference resolution, achieving high rankings on the SuperGLUE leaderboard (see Fig.~\ref{fig7}).
\end{itemize}

\subsubsection{Financial Domain}
\label{financial_domain_challenges_and_key_findings}
\begin{itemize}
    \item The 50-billion parameter LLM excels in financial tasks like investment analysis, risk management, and fraud detection.
    \item FinBERT achieved SOTA results in financial sentiment analysis, particularly on the FiQA sentiment scoring and Financial PhraseBank datasets, with an F1 score of 0.95 on the Financial PhraseBank dataset.

\end{itemize}

\subsubsection{Nuclear Domain}
\label{nuclear_domain_challenges_and_key_findings}
\begin{itemize}
    \item NukeBERT, fine-tuned on nuclear-specific datasets (NText and NQuAD), achieved significant improvements over BERT, with F1 and exact match scores of 93.87 and 88.31.
    \item NUKELM, trained on a large dataset from the OSTI database, excelled in classifying research articles and outperformed NukeBERT in binary classification tasks.

\end{itemize}

\subsubsection{Humanitarian Domain}
\label{humanitarian_domain_challenges_and_key_findings}
\begin{itemize}
    \item HumBERT, trained on humanitarian text sources, outperformed generic language models in zero-shot and full-training scenarios, achieving an average F1 task score of 0.73.
    \item The HUMSETBIAS dataset was introduced to assess and mitigate biases related to gender and country, effectively reducing biases without compromising performance.

\end{itemize}

\subsubsection{Social Science Domain}
\label{social_science_domain_challenges_and_key_findings}
\begin{itemize}
    \item SsciBERT, designed for social science texts and trained on social science abstracts, outperformed baseline models in classification, identification, and recognition tasks.
    \item The SSCISciBERT-e4 variant achieved the highest weighted average F1 score of 60.75\%, representing a significant contribution to pre-trained models for social sciences.
\end{itemize}

\subsubsection{Material Domain}
\label{material_domain_challenges_and_key_findings}
\begin{itemize}
    \item MatSciBERT, trained on materials science literature, outperformed SciBERT in tasks like abstract classification, NER, and RE, with a high F1 score of 93.57\% in abstract classification.
    \item The model leverages knowledge from computer science and biomedical corpora, demonstrating superior performance and potential for information extraction in materials science texts.
 
\end{itemize}

\emph{General Challenges Across Domains:}
\begin{itemize}
    \item Specialized Vocabulary and Complex Structures: Domain-specific texts often contain specialized terminology and complex linguistic features, posing challenges for traditional NLP methods.
    \item Domain Adaptation and Generalization: Adapting PLMs to specific domains requires extensive pre-training on relevant corpora and ensuring the models can generalize to new and evolving topics.
    \item High Computational Costs: Training and fine-tuning domain-specific models demand significant computational resources, which can be a barrier to adoption and deployment.
    \item Ethical Considerations: The deployment of PLMs raises ethical concerns, including data privacy, bias in model outputs, and the need for transparency and interpretability.

\end{itemize}

\emph{Specific Challenges by Domain:}
\begin{itemize}
    \item Biomedical:
    \begin{itemize}
        \item Variability in Task Performance: PLMs exhibit varying performance across different biomedical tasks, requiring task-specific fine-tuning.
    \end{itemize}
    \item Computer Science and Biomedical: 
    \begin{itemize}
        \item High Cost of Annotated Data: Obtaining large amounts of labeled data is expensive and labor-intensive, especially in scientific fields. 
        \item Balancing Generalization and Specialization: Ensuring models perform well across various scientific domains and tasks.
    \end{itemize} 
    \item Financial:
    \begin{itemize}
        \item Built-in Biases and Data Volume: Managing inherent biases and processing vast amounts of unstructured data effectively.
    \end{itemize}
    \item Nuclear:
    \begin{itemize}
        \item Limited Public Data and Data Optimization: Addressing the scarcity of publicly available data and developing efficient data optimization techniques.
    \end{itemize}
    \item Humanitarian:
    \begin{itemize}
        \item Bias in Data and Resource Constraints: Mitigating intrinsic biases and ensuring access to computational resources and expertise.
        \item Generalization Across Tasks: Ensuring consistent performance across various humanitarian tasks.
    \end{itemize}
    \item Social Science:
    \begin{itemize}
        \item Limited Training Data and Generalization: Training on abstracts may limit generalization to full texts, and the absence of standardized datasets hinders robust evaluation.
        \item Performance Metrics: Improving the F1 score to comprehensively capture and understand social science texts.
    \end{itemize}
    \item Material:
    \begin{itemize}
        \item Dataset Size and Diversity: Ensuring a large and diverse dataset for training and fine-tuning.
        \item Evaluation Metrics: Developing comprehensive benchmarks and ensuring high performance across different tasks.
    \end{itemize}
By addressing these challenges, domain-specific PLMs can continue to advance and provide significant contributions to their respective fields.

\end{itemize}

\begin{figure}[h]
  \centering
  \includegraphics[width=\linewidth]{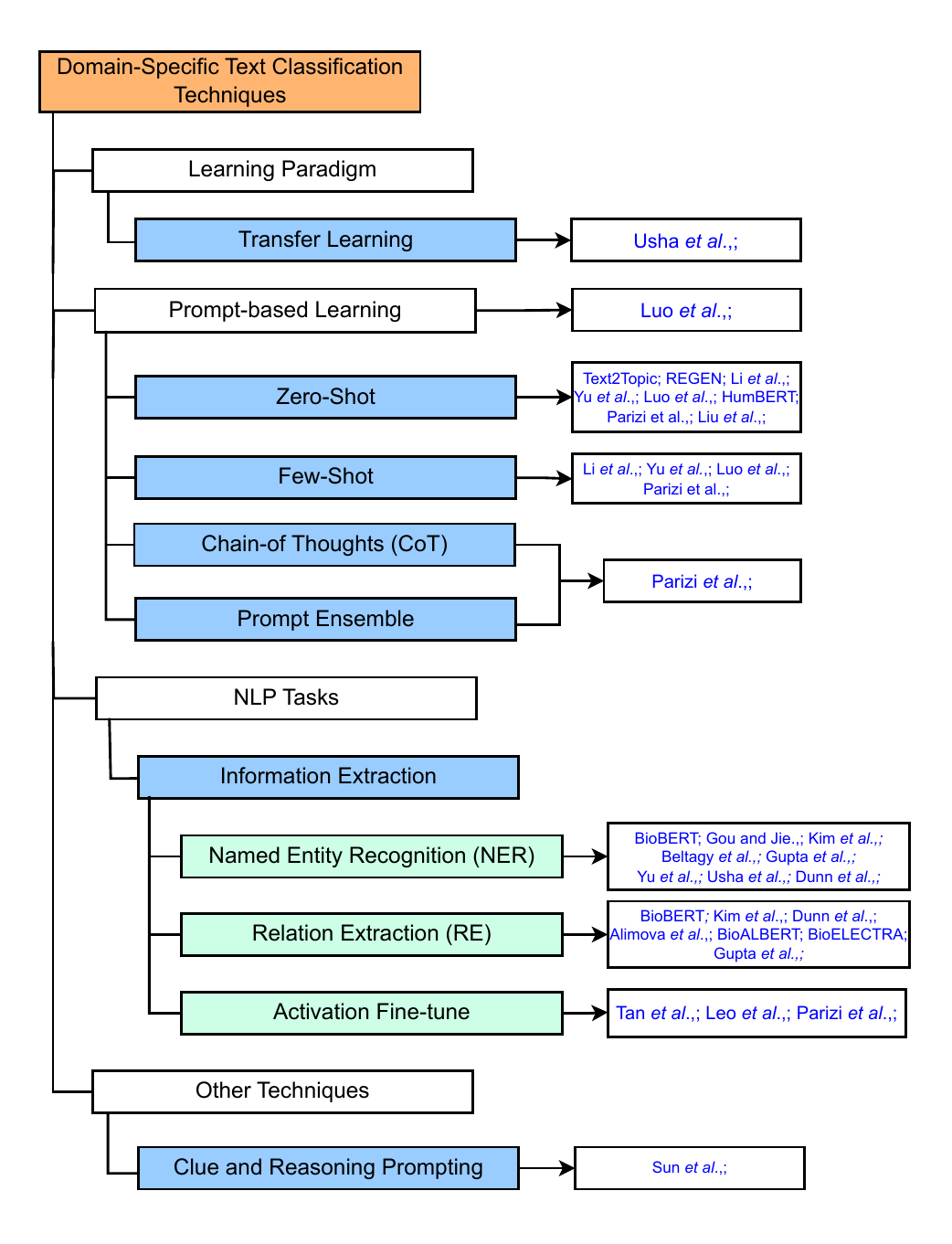}
  \caption{An overview of domain-specific text classification techniques.}
  \Description[Commonly utilized techniques for domain-specific text classification.]{The figure shows a taxonomy for recently utilized techniques for domain-specific text classification.}
  \label{fig8}
\end{figure}

\section{Domain-Specific Text Classification Techniques}
\label{sec:domain_specific_text_classification_techniques}
In this section, we will focus on the papers that investigate various learning techniques and their effectiveness in improving the performance of LLMs on text classification tasks (presented in Fig.~\ref{fig8} and Table~\ref{tab:sreviewed_studies_techniques}). 

\subsection{Learning Paradigm}
\label{subsubsec:learning_paradigm}
In this section we present various training techniques crucial for LLMs to achieve top performance. We delve into the specific methods employed by researchers. By applying these techniques strategically, researchers can push the boundaries of LLM capabilities and achieve SOTA results.

\subsubsection{Transfer Learning}
\label{subsubsec:transfer_learning}
Transfer learning is a fundamental technique where a PTM acts as a foundation. The PTM, already trained on a massive dataset, gets fine-tuned for a specific task using a smaller, more targeted dataset. This approach leverages the knowledge gained from the initial training to improve performance on the new, specialized task.

Neural network architecture has been utilized by Usha et al.~\cite{usha2022named} to propose an approach that utilized transfer learning with a combination of pre-trained SciBERT language model and bidirectional long short-term memory (Bi-LSTM) for BioNER. The advancement of biomedical research has led to an exponential growth of biomedical data. To effectively extract meaningful data from this amount of information, biomedical text mining and NLP have emerged as powerful tools. NER plays a crucial role in biomedical text mining, particularly in extracting data from scientific literature. Biomedical named entities are more challenging to identify due to their distinctive nomenclatural characteristics. SciBERT provides a higher efficiency while working with biomedical tasks, and Bi-LSTM network allows the collection of more comprehensive context data, for these reasons the model employed these techniques. The study aims to reduce the shortage of high quality, extensive scientific data through employed SciBERT, and effectively mitigate the excessive reliance on tags in sequence labeling tasks by applying a combination of Bi-LSTM and CRF models. The model utilizes a combination of a pre-trained SciBERT language model and a Bi-LSTM layer to achieve superior performance. SciBERT (stated in section~\ref{subsubsec:pre_trained_model_for_computer_science_and _tasks}), a language model specifically trained on scientific literature, provides rich contextual information and adaptability to biomedical terms, while Bi-LSTM effectively captures both past and future context for better NER. A CRF layer is additionally employed to handle the dependencies between tags in sequence labeling. Evaluation of the NCBIDisease corpus demonstrates the effectiveness of the proposed model, achieving a high F1 score of 91.00\%, outperforming existing methods.

\subsubsection{Learning Paradigm Key Findings and Challenges}
\label{subsubsec:learning_paradigm_key_findings_and_challenges}
This section delves into the key findings and challenges in the learning paradigm for domain-specific text classification.

\emph{Key Findings:}
\begin{itemize}
    \item Transfer Learning:
    \begin{itemize}
        \item Uses a PTM fine-tuned on a smaller, task-specific dataset to enhance performance on specialized tasks.
        \item BioNER combined SciBERT and Bi-LSTM for biomedical named entity recognition (BioNER), improving data extraction from biomedical literature.
    \end{itemize}
    \item Model Components:
    \begin{itemize}
        \item SciBERT: Provides rich contextual information and adaptability to biomedical terms.
        \item Bi-LSTM: Captures both past and future contexts.
        \item CRF: Manages dependencies between tags in sequence labeling, enhancing accuracy.
    \end{itemize}
    \item Performance:
    \begin{itemize}
        \item The model achieved a high F1 score of 91.00\% on the NCBIDisease corpus, outperforming existing methods.
    \end{itemize}

\end{itemize}

\emph{Challenges:}
\begin{itemize}
    \item Domain-Specificity:
    \begin{itemize}
        \item Biomedical Text Complexity: Biomedical entities are harder to identify due to their unique characteristics.
        \item Data Availability: There's a lack of high-quality, extensive scientific data.
    \end{itemize}	
    \item Model Integration:
    \begin{itemize}
        \item Combining Techniques: Balancing multiple techniques (SciBERT, Bi-LSTM, CRF) without adding excessive complexity.
    \end{itemize}
    \item Computational Resources:
    \begin{itemize}
        \item High Requirements: Significant computational resources are needed for training and fine-tuning large models.
    \end{itemize}

\end{itemize}

\subsection{Prompt-based Learning}
\label{subsec:prompt_based_learning}
This technique offers a unique way to train LLMs.  Instead of relying only on massive datasets, this approach utilizes carefully crafted instructions called prompts. These prompts essentially act like a guidance for the LLM.  By providing specific context and outlining the desired outcome, prompts guide the LLM towards successfully completing a task. This allows researchers to tailor tasks to the LLM's strengths and achieve better results.

Prompt-learning has been employed by Luo et al.~\cite{luo2023exploring} to overcome the problem of the significant amount of labeled data that is required during the fine-tuning phase, especially in a domain-specific application. Prompt-learning is known for its efficiency within few-shot scenarios. In addition to it, small language models (SLM: models with less than 1B parameters) provide significant customizability, adaptability, and cost-effectiveness for domain-specific tasks. The study examined the possibility of leveraging SLMs in combination with the prompt-learning approach to enhance domain-specific text classification, particularly within customer-agent conversations in the retail sector. In the case of few-shot settings, while prompt-learning approach was used with an abomination of SLMs for domain-specific text classification, the model attained 75\% accuracy with 15\% of full labeled data utilizing T5-base, these achievements are comparable with the traditional fine-tuning approach. The model significantly boosts performance after validating the active few-shot sampling and ensemble strategies in the prompt-learning process. In a zero-shot scenario, even though the massive GPT-3.5-turbo with approximately 154B parameters achieves a remarkable 55.16\% accuracy, these optimized prompts allow for significantly smaller models like FLAN-T5-large to achieve over 30\% accuracy, surpassing over half the performance of GPT-3.5-turbo while maintaining a significantly smaller parameter footprint. The research highlights the effectiveness of prompt-learning techniques in domain-specific text classification, particularly for SLMs. Few-shot prompt-based model fine-tuning, active few-shot sampling, and ensemble strategies contribute to improved performance. Detailed instructions in zero-shot settings further boost the model's capabilities.

\begin{table}
  \caption{Reviewed Studies: Techniques Utilized for Domain-Specific Text Classification.}
  \label{tab:sreviewed_studies_techniques}
   \resizebox{\textwidth}{!}{%
  \begin{tabular}{|c|c|cccc|cc|c|c|}
    \toprule

\multicolumn{1}{|c|}{\multirow{2}{*}{Works}} & \multicolumn{1}{|c|}{Learning Paradigm} & \multicolumn{4}{|c|}{Prompt-based Learning}
& \multicolumn{2}{|c|}{Information Extraction}
& \multicolumn{1}{|c|}{\multirow{2}{*}{Activation Fine-tuning}} & \multirow{2}{*}{CARP}
\\ 
\cline{2-8}
\multicolumn{1}{|c|}{}
& \multicolumn{1}{|c|}{Transfer Learning}
& \multicolumn{1}{c}{Zero-Shot}
& \multicolumn{1}{c}{Few-Shot}
& \multicolumn{1}{c}{Chain-of-Thought}
& \multicolumn{1}{c}{Ensemble}
& \multicolumn{1}{|c}{NER}
& \multicolumn{1}{c|}{RE}
& 
& \multicolumn{1}{|c|}{}
\\ \cline{1-10}

Usha et al.~\cite{usha2022named}
& \checkmark
&
&
& 
& 
&
\checkmark
&
&
&
\\
\hline
Luo et al.~\cite{luo2022biogpt}
&
&
\checkmark
&
\checkmark
& 
& 
\checkmark
& 
&
&
\checkmark 
&
\\
\hline
Wang et al.~\cite{wang2023text2topic}
&
&
\checkmark
&
&
&
&
&
&
&
\\
\hline
REGEN~\cite{yu2023regen}
&
&
\checkmark
&
&
& 
&
& 
&
& 
\\
\hline
Li et al.~\cite{li2023synthetic}
&
&
\checkmark
& 
\checkmark
& 
& 
&
& 
& 
& 
\\
\hline
Yu et al.~\cite{yu2023open} 
&
&
\checkmark
&
\checkmark
&
&
&
\checkmark &
&
\checkmark
&
\\
\hline
Parizi et al.~\cite{parizi2023comparative} 
&
&
\checkmark
&
\checkmark
&
\checkmark
&
&
&
&
\checkmark
&     
\\
\hline
Dunn et al.~\cite{dunn2022structured}
&
&
&
&
&
&
\checkmark
&
\checkmark &
&
\\
\hline
Alimova et al.~\cite{alimova2021cross} 
&
&
&
&
&
&
&
\checkmark &
&
\\
\hline
Gou and Jie~\cite{gou2023lightweight}
&
&
&
&
&
&
\checkmark &
&
&
\\
\hline
Sun et al.~\cite{sun2023text}
&
&
&
&
&
&
&
&
&
\checkmark 
\\
\hline
\end{tabular}%
}
\end{table}

\subsubsection{Zero-Shot and Few-Shot}
\label{subsubsec:zero_and_few_shot}
Zero-shot and few-shot are learning approaches for training language models on new tasks using zero or a small number of training examples. These techniques have several advantages over traditional fine-tuning, including the need for less task-specific training data and a reduced potential to overfit to the training data. However, fine-tuning may still outperform these approaches in some cases~\cite{brown2020language}. 

Wang et al.~\cite{wang2023text2topic} use a Bi-Encoder Transformer architecture for achieving high multi-label classification performance. The outcome of the study is a flexible framework known as Text2Topic that efficiently classifies text into multiple topics. The model architecture utilizes concatenation, subtraction, and multiplication of embeddings on both text and topic. The model architecture also supports both zero-shot capabilities and exhibits domain-specific text embeddings. The study developed a Text2Topic model that classifies text into multiple topics. The model was trained on a large-scale dataset of text-topic pairs collected from Booking.com using optimized sampling and labeling methods. The model achieved superior performance compared to other models, with micro and macro mAP scores of 92.9\% and 75.8\%, respectively.

REGEN (Retrieval-Enhanced Zero-Shot Data Generation) has been proposed by Yu et al.~\cite{yu2023regen} as a novel method for zero-shot text classification. REGEN's retrieval-enhanced framework effectively utilizes a general-domain unlabeled corpus to construct high-quality training data, significantly reducing the reliance on hand-labeled data. To further enhance the quality of synthetic data, REGEN employs two complementary strategies: verbalizer augmentation with demonstration and self-consistency guided filtering. Experiments conducted on nine benchmark NLP classification tasks demonstrate the remarkable effectiveness of REGEN, achieving a significant 4.3\% improvement over the strongest baselines while requiring 70\% less time to generate the training data compared to baselines using NLG models. REGEN seamlessly integrates with recently proposed language models, paving the way for further performance enhancements. High-quality training data is essential for classifying text accurately with high performance. Collecting and curation of these kinds of data is expensive with high demand of time. Currently, some studies are trying to use LLMs for building synthetic datasets to overcome the mentioned problems.

Authors from~\cite{li2023synthetic} employ two different approaches for synthetic data generation which are zero-shot, and few-shot settings. In the zero-shot settings, the LLM is directly instructed to produce text examples with various labels of interest, and in a few-shot method, a few genuine data samples are supplied as models to guide the LLM in generating the synthetic data. Two evaluation studies have been investigated, each focusing on one aspect of subjectivity: the first study assesses the effectiveness of synthetic data across 10 different types of classification tasks and examines how it varies depending on the task's level of subjectivity. The second study investigates how the performance of a model trained on synthetic data changes for a specific classification task when the subjectivity of the task instances varies. The study concluded that, among all the 10 various tasks experimented with in the study, the models trained on the real-world data outperformed those trained only within the LLM-generated synthetic data, also provided a few real-world data to guide LLM during the synthetic data generation process with positive impact of the effectiveness of the generated data. Moreover, the study discovered that the models trained on LLM-generated synthetic data perform nearly as well as those trained on real-world data for tasks with low subjectivity, but the performance gap is significantly greater for tasks with high subjectivity. As a result, these findings suggest that LLM-generated synthetic data can be a valuable tool for training machine learning models, but its effectiveness is likely to vary depending on the subjectivity of the task. For tasks with low subjectivity, LLM-generated synthetic data can be a nearly perfect substitute for real-world data, but for tasks with high subjectivity, human annotators will still be necessary to provide high-quality training data. The researchers used two pre-trained LLMs, BERT~\cite{devlin2018bert} and RoBERTa~\cite{liu2019roberta}, as encoders. The classification model itself comprised a hidden layer of 768 units and an output layer. The datasets have been randomly divided as follows: training 70\%, validation 5\%, and test 25\%. The model was fine-tuned with a learning rate of $5e\! - 5$ and a batch size of 64. The performance of the classification models was evaluated using Macro-F1 and accuracy scores. The experiments consistently indicated that models trained on real-world data generally outperform those trained on synthetic data. This is evident in the RoBERTa model, where the average improvements in Macro-F1 and accuracy for synthetic data were as follows: 16.9\% and 14.9\%, while trained with zero-shot respectively, and 6.7\% and 6.1\% improvement while trained with few-shot, respectively. Similarly, the BERT model also exhibits a similar pattern, showcasing the superiority of real-world data for training classification models. Based on the results obtained, the researchers found that utilizing real-world data examples during the LLM training process can considerably improve the efficacy of synthetic data. Moreover, they observed that models trained on synthetic data generated under the few-shot settings consistently outperformed their counterparts trained on synthetic data generated under the zero-shot settings. For instance, the BERT model trained on synthetic data generated under the few-shot setting achieved an average improvement of 10.6\% in terms of Macro-F1 and 8.8\% in terms of accuracy compared to the model trained on synthetic data generated under the zero-shot setting across the 10 tasks. Similarly, the RoBERTa model trained on synthetic data generated under the few-shot setting achieved an average improvement of 10.3\% in terms of Macro-F1 and 8.9\% in terms of accuracy compared to its counterpart trained on synthetic data generated under the zero-shot setting. These findings indicate that providing LLMs with real-world data examples can substantially enhance the quality and effectiveness of synthetic data for training classification models.

Yu et al.~\cite{yu2023open} have assessed the effectiveness of various LLMs across three NLP tasks: NER, political party prediction, and misinformation detection. The researchers compared closed generative LLMs (GPT-3.5 and GPT-4)~\cite{zhao2023survey}, open generative LLMs (Llama 2 13B and 70B)~\cite{touvron2023llama}, and a smaller non-generative LLM (RoBERTa) on eight datasets. They investigated the impact of various prompts and training setups, including zero-shot, few-shot, and fine-tuning. For NER classification, RoBERTa demonstrated superior performance on the CoNLL 2003 and WiKiNERT-EN datasets, with scores of 94.3 \textpm 3.5 and 96.2 \textpm 0.1, respectively. GPT-4 excelled in NER on the WNUT 2017 dataset, achieving a score of 65.1 \textpm 3.0. Regarding explicit ideology prediction, GPT-4 outperformed other models on both the 2020 Election and COVID-19 datasets, with scores of 97.6 \textpm 0.5 and 95.1 \textpm 0.6, respectively. RoBERTa achieved the highest accuracy on the 2021 Election dataset, with a score of 95.2 \textpm 0.7. For implicit ideology prediction, RoBERTa consistently demonstrated the best results across all three datasets: 2020 Election (93.0 \textpm 0.2), COVID-19 (70.0 \textpm 2.7), and 2021 Election (82.3 \textpm 1.1). In the realm of misinformation detection, GPT-3.5 emerged as the top performer on both the LIRA and CT-FAN-22 datasets, with scores of 68.5 \textpm 3.0 and 43.7 \textpm 1.9, respectively. The study revealed that supervised LLMs like RoBERTa, despite their smaller size, often match or even exceed the performance of generative LLMs, while offering advantages in terms of cost, processing speed, and transparency. Furthermore, the study emphasized the essential role of prompt engineering, highlighting the substantial impact of prompt selection on a model's performance. Prompting methods that excel in zero-shot settings may not yield comparable results in few-shot settings, underscoring the intricacies of prompt design. Additionally, the study demonstrated the capability of open-source LLMs like Llama 2 to surpass closed-source models like GPT-3.5 through fine-tuning techniques, signifying the potential of collaborative open-source endeavors. It is noteworthy that supervised LLMs like RoBERTa excel in tasks with well-defined patterns, while generative models like GPT-3.5 perform better in tasks that demand generalization and transferability.

\subsubsection{Chain-of-Thought and Prompt Ensemble}
\label{subsubsec:chain_of_thought_and_prompt_ensemble}
Recent advancements in LLMs have transformed them into powerful tools for various NLP tasks. The impact of using different prompt engineering strategies for classifying legal text has been investigated by Parizi et al.~\cite{parizi2023comparative}. The experiments involve various LLMs and prompt techniques, including zero-shot and few-shot prompting (Section~\ref{subsubsec:zero_and_few_shot}), prompt ensemble, chain-of-thought, and activation fine-tuning. The study uncovers that working with legal documents presents unique challenges for LLMs, such as complex terminologies and lengthy documents, which impede the transferability of prompt techniques to this domain. Two distinct open pre-trained transformers (OPT) models are selected for conducting the experiments. These models are available in multiple sizes and possess capabilities comparable to GPT-3, while also offering open-source features. ECHR and SCOTUS are the two English datasets employed in the study. ECHR encompasses various tasks, including a binary classification task that determines whether a human rights article has been violated. The SCOUTS dataset involves a multi-class classification task that classifies supreme court case decisions into 14 distinct issue areas. Due to computational resource limitations, these datasets are down-sampled to 10\% and 5\%, respectively. The study uses GPT-4 to adjust the average token length of SCOTUS documents to make it acceptable by OPT. Models like HIER-BERT and Legal-BERT are used for comparison with the achieved results for ECHR and SCOTUS, respectively. The study found that providing more context with task demonstration causes remarkable improvements as observed while going from zero-shot to few-shot prompting. Although the OPT models demonstrate some improvement, their performance still falls short of the supervised approaches. This implies that the OPT models, despite their large size and general-purpose capabilities, face challenges in handling the specialized legal tasks examined in this study without the assistance of fine-tuning or domain-adaptation techniques. CoT prompting was found to be less effective than zero-shot prompting for OPT models due to poor-quality output generations. This suggests that CoT's benefits depend on the LLM's ability to generate relevant reasoning. The lack of code pre-training in OPT and reasoning hallucinations may also hinder CoT prompts' effectiveness in legal domain. Activation finetuning proved to be the most effective approach for improving performance in legal text classification tasks. When combined with a few-shot prompt, it achieved a significant boost in macro-F1 scores of 35 and 47 points for both ECHR and SCOTUS datasets. Among the pure prompting approaches, few-shot prompts with and without ensembling showed superior performance, while CoT prompting failed to deliver comparable results.

\subsubsection{Prompt-Based Learning Key Findings and Challenges}
\label{subsubsec:prompt_based_learning_key_findings_and_challenges}
In this section, we delve into the key findings and challenges associated with prompt-based learning in domain-specific text classification. This technique leverages carefully crafted prompts to guide LLMs, allowing for efficient performance in zero-shot and few-shot scenarios. We explore the effectiveness of prompt-based learning, including the use of CoT and prompt ensemble strategies, and examine the challenges posed by domain-specific complexities, data dependence, and resource constraints.

\emph{Key Findings:}
\begin{itemize}
    \item Prompt-Based Learning Efficiency:
    \begin{itemize}
        \item It utilizes carefully crafted prompts to guide LLMs, reducing the need for large labeled datasets and making it effective in few-shot scenarios.
        \item Luo et al.~\cite{luo2023exploring} used small language models (SLMs) with prompt-learning for domain-specific text classification in customer-agent conversations, achieving 75\% accuracy with only 15\% of the labeled data.
    \end{itemize}
    \item Performance in Zero-Shot and Few-Shot Settings:
    \begin{itemize}
        \item Zero-Shot: Smaller models like FLANT5-large achieved over 30\% accuracy with optimized prompts, surpassing half the performance of GPT-3.5-turbo but with fewer parameters.
        \item Few-Shot: Achieved superior performance through active few-shot sampling and ensemble strategies.
    \end{itemize}
    \item CoT and Prompt Ensemble:
    \begin{itemize}
        \item CoT: Found to be less effective than zero-shot prompting in legal domain tasks for some models, likely due to poor-quality output generations and lack of reasoning capabilities in the models.
    \end{itemize}
\end{itemize}

\emph{Challenges:}
\begin{itemize}

    \item Domain-Specific Complexity:
    \begin{itemize}
        \item Task Complexity: Legal and biomedical texts pose unique challenges due to complex terminologies and lengthy documents.
        \item Model Limitations: Large general-purpose models may underperform without fine-tuning or domain-specific adaptation.
    \end{itemize}
    \item Data Dependence:
    \begin{itemize}
        \item Synthetic Data Quality: Real-world data generally outperforms LLM-generated synthetic data, especially for high subjectivity tasks.
        \item Data Generation: High-quality training data is expensive and time-consuming to collect.
    \end{itemize}
    \item Resource Constraints:
    \begin{itemize}
        \item Computational Resources: Training and fine-tuning large models require significant computational power, making it less feasible for smaller teams or institutions.
    \end{itemize}
\end{itemize}

\subsection{NLP Tasks}
\label{subsec:NLP_tasks}
Another category in domain-specific text classification techniques pertains to NLP tasks, which involve the computational analysis, understanding, and generation of human languages. In this section, we will explore some commonly used techniques for domain-specific text classification, highlighting their applications and effectiveness within specialized fields.

\subsubsection{Information Extraction}
\label{subsubsec:information_extraction}
Information extraction is a specific type of NLP task focused on gathering specific details from text. The extracted information can then be organized, categorized, and stored in a structured format, making it easier to analyze and use for various applications

\paragraph{Named Entity Recognition and Relation Extraction}
\label{paragraph:named_entity_recogntion_and_relation_extraction}
A large portion of scientific knowledge about solid-state materials is dispersed across an extensive collection of academic publications, existing in the form of text, tables, and figures. To extract meaningful and structured information from this complex scientific context, a sequence-to-sequence approach has been proposed by~\cite{dunn2022structured}. This approach combines NER and RE~\cite{zhao2023comprehensive} to identify and extract relevant entities and their relationships. Specifically, GPT-3, a large language model, is fine-tuned on around 500 document-completion examples to achieve the desired structured output, such as lists of JSON documents. Flexibility and accessibility are two advantages of the approached model besides its high level of accuracy. Almost any information extraction task can be addressed by defining a new output format, and scientific experts can readily create their own models by carefully reading relevant passages and documenting the desired output structure. The study achieved excellent performance on three different applications for material engineering: solid-state impurity doping, metal-organic frameworks (MOF), and general materials relations. Understanding the connections between elements within a text is a crucial step in downstream tasks, whether it be supervised machine learning or the development of knowledge graphs. The RE~\cite{zhao2023comprehensive} models utilize a predefined set of relations to determine which entities are linked. The study utilizing 100 – 1000 manually annotated text-extraction (prompt(input) – completion (output)) pairs to apply NERRE tasks by the fine-tuning model based on GPT-3 (davinci, 175B parameters) demonstrates impressive performance on three distinct tasks: dopant-host material linking, MOF cataloging, and general chemistry information extraction. The achieved extract matches F1 scores of 94\% (doping-ENG: host), 71.8\% (MOF-JSON: name), and 71.5\% (general-JSON: applications), respectively, outshine previous methods and showcase the versatility of the proposed approach. 

Alimova et al.~\cite{alimova2021cross} conduct a comprehensive evaluation of RE methods across scientific abstracts and electronic health records within the biomedical field. It specifically focuses on evaluating the effectiveness of advanced neural architectures, including LSTM with cross-attention and BERT. The evaluation involves comparing the performance of these models on four biomedical datasets: PHAEDRA, i2b2/VA, BC5CDR, and MADE corpora. The findings reveal that while BioBERT~\cite{lee2020biobert} experiences a significant decrease in performance (F1-measure of 34.2\%) when applied to out-of-domain data, the cross-attention LSTM model performs better with a smaller decline in accuracy (F1-measure 27.6\%). This highlights the importance of choosing appropriate neural architectures for biomedical relation extraction tasks, particularly when dealing with diverse datasets.

Gou and Jie~\cite{gou2023lightweight} introduce LWNER, a novel lightweight BioNER method based on the BERT model~\cite{devlin2018bert}. LWNER leverages the Transformer architecture to encode contextual information from large text corpora, eliminating the need for extensive manual feature engineering. LWNER employs a deep learning architecture, including token and segment embedding, Convolutional NN, Transformer encoder, BiLSTM, and CRF blocks, to efficiently extract information. Fine-tuning BERT on biomedical data further enhances its ability to recognize entities specific to the domain. The authors demonstrate the effectiveness of LWNER on various BioCreative tasks, achieving an F1 score of 91.3\% for chemical entity recognition in the BC5CDR dataset. Additionally, they present an online tool built upon LWNER, enabling researchers to extract valuable information from scientific literature and facilitate knowledge graph construction.

\subsubsection{Activation Fine-tuning}
\label{subsubsec:activation_fine_tuning}
Parizi et al.~\cite{parizi2023comparative} (Section~\ref{subsubsec:chain_of_thought_and_prompt_ensemble}) investigated various methods to enhance the performance of LLMs in classifying legal documents. While alternative prompt-based strategies like zero-shot and few-shot prompting showed some improvement, activation fine-tuning emerged as the most effective approach. This method involves refining specific components of a LLM tailored to the task of legal text classification. By combining activation fine-tuning with a few-shot prompt, the study achieved significant enhancements in accuracy across both datasets utilized for legal judgments (ECHR and SCOTUS). These findings suggest that for intricate legal tasks, directly fine-tuning the LLM yields greater benefits compared to relying only on prompt engineering methods.

\subsubsection{NLP Tasks Key Findings and Challenges}
\label{subsubsec:nlp_tasks_key_findings_and_challenges}
In this section, we delve into the key findings and challenges associated with NLP tasks in domain-specific text classification. These tasks encompass various techniques for the computational analysis, understanding, and generation of human languages, with a focus on their applications and effectiveness in specialized fields.

\emph{Key Findings:}
\begin{itemize}
    \item Information Extraction through NER and IE:
    \begin{itemize}
        \item Combining NER and RE using a sequence-to-sequence approach, fine-tuned GPT-3 models can achieve high accuracy in extracting structured information from scientific texts.
        \item This approach demonstrated excellent performance in three applications within material engineering, achieving high F1 scores in tasks like dopant-host material linking, MOF cataloging, and general chemistry information extraction.
        \item For biomedical contexts, models like BioBERT and LSTM with cross-attention exhibit varying performance based on dataset specificity, highlighting the importance of model selection for different biomedical datasets.
        \item Lightweight models like LWNER, built on the BERT architecture, effectively recognize entities in biomedical data, achieving an F1 score of 91.3\% in chemical entity recognition.
    \end{itemize}
    \item Activation Fine-tuning:
    \begin{itemize}
        \item Activation fine-tuning has been shown to significantly enhance the performance of LLMs in legal text classification.
        \item When combined with few-shot prompting, activation fine-tuning yielded substantial improvements in accuracy for legal judgment datasets (ECHR and SCOTUS), outperforming other prompt-based strategies.
    \end{itemize}
\end{itemize}

\emph{Challenges:}
\begin{itemize}
    \item Data Specificity and Generalization:
    \begin{itemize}
        \item NLP models often face challenges when applied to out-of-domain data, as evidenced by the significant performance drops seen in models like BioBERT when used on different biomedical datasets.
        \item Ensuring models can generalize across different datasets and domains remains a critical challenge.
       
    \end{itemize}
    \item  Model Complexity and Resources:
    \begin{itemize}
        \item While large models like GPT-3 demonstrate high performance, their size and resource requirements can be prohibitive. Lightweight models offer a more resource-efficient alternative but may require careful tuning to match the performance of larger models.
        \item Activation fine-tuning, although effective, demands substantial computational resources and expertise to implement properly.
    \end{itemize}
    \item Task-Specific Adaptation:
    \begin{itemize}
        \item Developing models that can handle the complexities of domain-specific tasks, such as legal text classification, requires specialized adaptation techniques beyond standard prompt engineering.
        
        \item For tasks with complex terminologies and structured information, models must be carefully fine-tuned to achieve optimal performance.
    \end{itemize}
\end{itemize}

While domain-specific NLP tasks show great promise in enhancing text classification through techniques like NER, RE, and activation fine-tuning, they also present significant challenges related to data specificity, resource demands, and the need for specialized adaptation.

\subsection{Other Techniques}
\label{subsec:other_techniques}

\subsubsection{Clue and Reasoning Prompting}
\label{subsubsec:clue_and_reasoning_prompting}
Linguistic reasoning is crucial for accurately classifying text based on its meaning and structure, Sun et al.~\cite{sun2023text} have proposed adopting a progressive reasoning approach to overcome complex linguistic phenomena that models struggle with. Clue And Reasoning Prompting (CARP) is a strategy that was introduced by the study, in the first step figures out superficial clues (such as keywords, tones, semantic relations, references, etc) in the received text, then CARP utilizes the clues and input as the basis for a diagnostic reasoning process, finally the decision depends on the two previous steps. As a result, the progressive reasoning approach effectively impacts on improving LLM’s ability in language reasoning contributing to text classification. The model utilizes an effective search to address the limited tokens allowed in context. CARP leverages a fine-tuned model trained on a labeled dataset to conduct kNN demonstration search for in-context learning. Across five widely used text classification benchmarks CARP outperformed on four out of five benchmarks: 97.39 (+1.24) on SST-2, 96.40 (+0.72) on AGNews, 98.78 (+0.25) on R8 and 96.95 (+0.6) on R52 and CARP achieved a performance of 92.39\%, which is only 0.91\% less than the SOTA on MR of 93.3\%. Also, it requires fewer examples per class to achieve comparable performance to supervised models with 1,024 examples per class. This highlights CARP's remarkable ability to excel in scenarios with limited training data.

\subsubsection{CARP Key Findings and Challenges}
\label{subsubsec:carp_key_findings_and_challenges}
In this section, we delve into the CARP technique for domain-specific text classification, examining its key findings and challenges.

\emph{Key Findings:}
\begin{itemize}
    \item Enhanced Linguistic Reasoning: CARP improves text classification by using superficial clues (keywords, tones, semantic relations) in a diagnostic reasoning process. CARP demonstrated superior performance on four out of five benchmarks, including 97.39\% on SST-2 and 98.78\% on R8.
    \item Data Efficiency: CARP achieves high accuracy with fewer training examples compared to traditional supervised models, proving effective even with limited training data.
\end{itemize}

\emph{Challenges:}
\begin{itemize}
    \item Handling Complexity: Accurately interpreting complex linguistic phenomena remains challenging, requiring sophisticated reasoning capabilities.
    \item Context Limitations: Managing the limited tokens allowed in context efficiently is difficult, especially for longer or more complex texts.
    \item Achieving SOTA Performance: While CARP performs well, consistently reaching SOTA performance is an ongoing challenge, requiring continuous refinement.
\end{itemize}

\subsection{Comparative Studies for Text Classification using LLMs}
\label{subsec:comparative_studies_for_text_classification_using_llms}
In this section, we present a comparative study that assesses the feasibility of distinguishing between summaries generated by ChatGPT and human-written summaries using text classification algorithms. Soni et al.~\cite{soni2023comparing} evaluate the quality and performance of abstractive summarization generated by ChatGPT using both human evaluation and automated metrics. A dataset of 50 summaries generated by ChatGPT was created. The summaries were scored using various automated metrics and evaluated by blinded human reviewers. Finally, a text classifier was built to distinguish between real and generated summaries. The real dataset sources from 50 articles from CNN News/Daily News dataset. It utilizes version 3.0.0 of the test-set. The dataset contains 287,226 training pairs, 13,368 validation pairs, and 11,490 test pairs. Also, 50 articles were utilized to generate summaries from CNN News/Daily News Dataset by prompting ChatGPT. The study utilizes various automated metrics to compare original and generated summaries. It employs ROUGE to calculate ROUGE-1, ROUGE-2, and ROUGE-L F1 scores, and METEOR to measure word-overlap, bigram-overlap, longest common sequence, and unigram-overlap between the real and generated summaries. The scores are obtained using the Hugging Face evaluation library. The study achieved 0.30, 0.11, 0.20, 0.21, and 0.35 for ROUGE-1, ROUGE-2, ROUGE-L, ROUGELSUM, and METEOR between real and generated summaries, respectively. Also, it was noticed through the reviewer’s feedback that they were unable to differentiate between original and generated human summaries, the accuracy for this task was 0.49. The study experimented DistilBERT and fine-tuned on the dataset, the model achieved 90\% in identifying ChatGPT generated summaries. The study presents that algorithms can distinguish between original and generated summaries that were built by ChatGPT or a human.

\section{Experiments}
\label{expermints}
In this section, we aim to present an evaluation of the effectiveness of domain-specific against general-purpose PLMs for tasks of sentence sequencing in biomedical research text comprehension. As a case study, we compare BERT (a general-purpose PLMs)~\cite{devlin2018bert} with BioBERT~\cite{lee2020biobert} and SciBERT~\cite{beltagy2019scibert} both of which are pre-trained on domain-specific corpora: biomedical literature and scientific text, respectively (see Sections~\ref{subsubsec:pre_trained_model_for_biomedical_tasks}, and~\ref{subsubsec:pre_trained_model_for_computer_science_and _tasks}). The objective is to evaluate the impact of domain adaptation on the models' performance in text classification. 

\subsection{Dataset}
The dataset utilized in this experiment is a PubMed 20k which is a subset of the larger PubMed 200 RCT collection prepared by Dernoncourt and Lee~\cite{dernoncourt-lee-2017-pubmed}, specifically designed for the sequence classification task in the medical literature. The dataset focuses on abstracts from the randomized controlled trial (RCT). The PubMed 20k is composed of the abstracts of RCTs taken from PubMed, and each sentence is labeled based on its role in the structure according to one of five categories: Background, Objective, Method, Result, or Conclusion. With automated sentence extraction this corpus supports research into a clinical abstract role assignment, aiding medical literature information retrieval. The dataset includes 15,000 abstracts for training, 2,500 for validation, and another 2,500 for testing, with a total of 180,040, 30,212, and 30,135 sentences respectively (see Table~\ref{tab:pubmed20k_stats}).

\begin{table}[ht]
\centering
\caption{Overview of the PubMed 20k dataset.}
\label{tab:pubmed20k_stats}
\begin{tabular}{lcc}
\toprule
\textbf{Split} & \textbf{Abstracts} & \textbf{Sentences} \\
\midrule
Training       & 15,000                       & 180,040 \\
Validation     & 2,500                        & 30,212  \\
Test           & 2,500                        & 30,135  \\
\bottomrule
\end{tabular}
\end{table}

\subsection{Experimental Design}
In the experimental setup, we follow a typical fine-tuning approach utilizing train, validation, and test splits provided with the PubMed 20k RCT dataset. Prior to feeding the texts to the models, the data is cleaned and preprocessed to avoid empty or malformed entries. Sentence labels are transformed into numerical representations.
Sentences are tokenized using the pre-trained tokenizer associated with each model. Tokens are restricted to 128, with appropriate truncation and padding applied.

Each model is fine-tuned as a sentence classifier with five output categories using the AdamW optimizer and a linear learning rate scheduler. To explore sensitivity to optimization parameters, training is repeated with four different learning rates, running for a maximum of 20 epochs in each case.  Early stopping was triggered if validation performance, measured by the micro F1 score, did not improve for 7 consecutive epochs (see Table~\ref{tab:training-setup}).

The best saved model that shows the highest validation F1 score during training is used to check each model's performance on the test set. We focus on precision, recall, and F1 score evaluation after completion.

\begin{table}[ht]
\centering
\caption{Fine-tuning Setup and Hyperparameters}
\label{tab:training-setup}
\begin{tabular}{ll}
\toprule
\textbf{Parameter} & \textbf{Value} \\
\midrule
Pretrained Models Used       & BERT-base, SciBERT, BioBERT \\
Sequence Length              & 128 tokens \\
Batch Size                   & 32 \\
Optimiser                    & AdamW \\
Learning Rates Tested        & 2e-5, 5e-6, 1e-6, 2e-6 \\
Learning Rate Scheduler      & Linear with warm-up \\
Warm-up Steps                & 10\% of total steps \\
Max Epochs per LR Schedule   & 20 \\
Early Stopping Patience      & 7 epochs \\
Loss Function                & Cross-Entropy Loss \\
Evaluation Metric            & Micro-averaged F1 score \\
Tokenizer                    & Model-specific (e.g., SciBERT tokenizer) \\
Hardware                     & NVIDIA GPU (Colab environment) \\
\bottomrule
\end{tabular}
\end{table}

\subsection{Results and Analysis}
We assess model performance using standard metrics including accuracy, precision, recall, and per-category F1 score. The results highlighted the influence of domain-specific pre-training on classification performance across different discourse roles in scientific text.

\subsubsection{Performance Analysis}
\begin{figure}[h]
  \centering
  \includegraphics[width=0.9\linewidth]{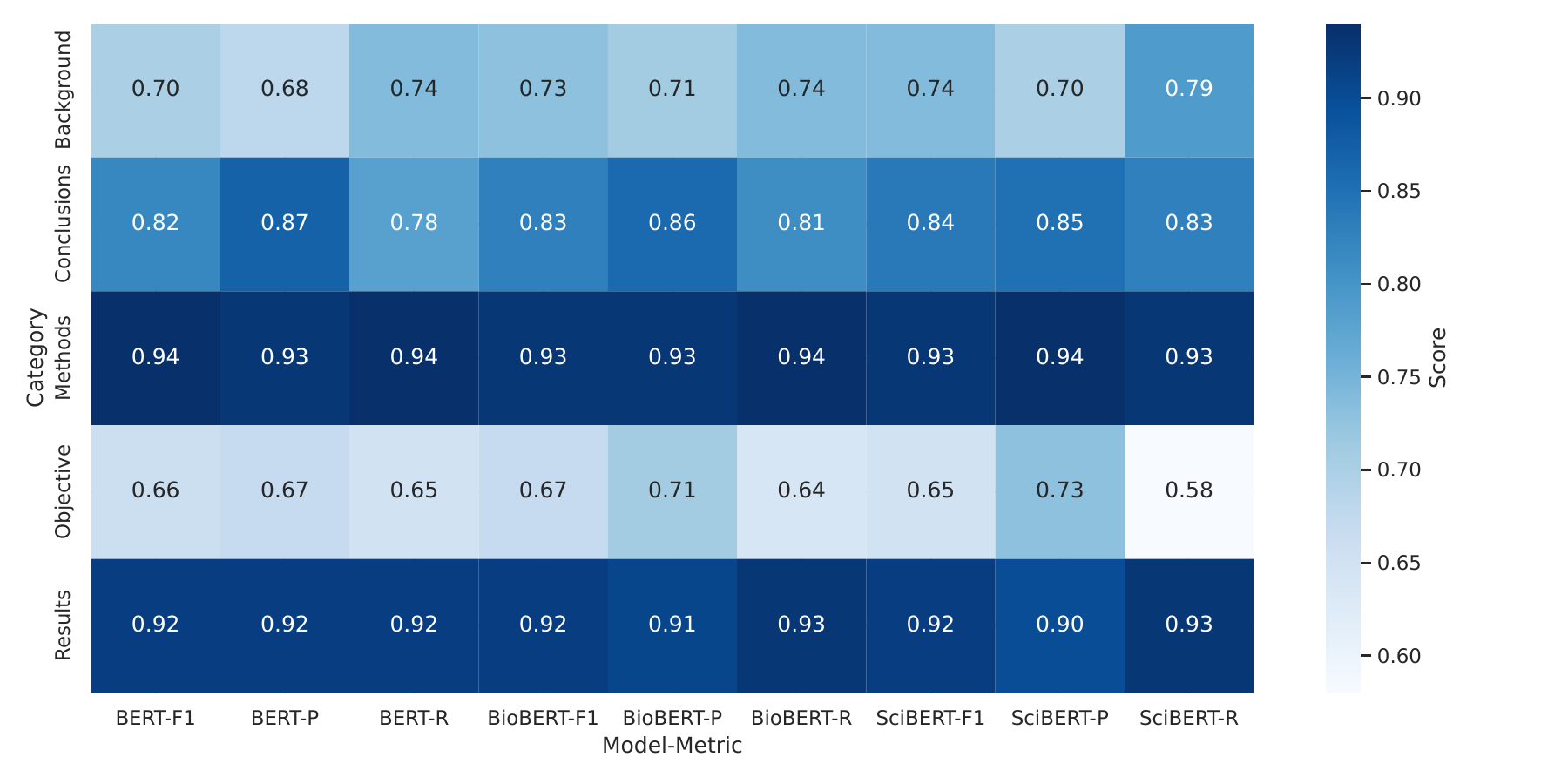}
  \caption{Heatmap of Precision(P), Recall(R), and F1 score(F1) per-category across BERT, BioBERT, and SciBERT.}
  \Description[Performance analysis per-category and model]{This figure presents Precision(P), Recall(R), and F1 score(F1) for individual class and model.}
  \label{fig9}
\end{figure}

Across investigated PLMs, \textit{Results} and \textit{Methods} are the easiest category combinations to predict, achieving high F1 scores ($\geq$ 0.92). As for \textit{Objective} sentences, they remain the most challenging, possibly owing to brevity and contextuality effect. Background and conclusions benefit most from domain-specific pre-training, as evidenced by SciBERT's superior performance (Fig.~\ref{fig9} presents a heatmap of precision (P), recall (R), F1 scores (F1) across categories and models). 

BERT demonstrates good precision for the categories \textit{Methods} and \textit{Results}, but lower recall for the categories \textit{Objective} and \textit{Conclusions}.

\begin{table}[ht]
\centering
\caption{Precision, Recall, and F1 score Per-Category for BERT, BioBERT, and SciBERT}
\label{tab:per_class_metrics}
\resizebox{\textwidth}{!}{
\begin{tabular}{lcccccccccc}
\toprule
\multirow{2}{*}{\textbf{Class}} & \multicolumn{3}{c}{\textbf{BERT}} & \multicolumn{3}{c}{\textbf{BioBERT}} & \multicolumn{3}{c}{\textbf{SciBERT}} \\
\cmidrule(lr){2-4} \cmidrule(lr){5-7} \cmidrule(lr){8-10}
 & Precision & Recall & F1 score & Precision & Recall & F1 score & Precision & Recall & F1 score \\
\midrule
Background   & 0.68 & 0.74 & 0.70 & 0.71 & 0.74 & 0.73 & 0.70 & 0.79 & \textbf{0.74} \\
Conclusions  & 0.87 & 0.78 & 0.82 & 0.86 & 0.81 & 0.83 & 0.85 & 0.83 & \textbf{0.84} \\
Methods      & 0.93 & 0.94 & \textbf{0.94} & 0.93 & 0.94 & 0.93 & 0.94 & 0.93 & 0.93 \\
Objective    & 0.67 & 0.65 & 0.66 & 0.71 & 0.64 & \textbf{0.67} & 0.73 & 0.58 & 0.65 \\
Results      & 0.92 & 0.92 & \textbf{0.92} & 0.91 & 0.93 & \textbf{0.92} & 0.90 & 0.93 & \textbf{0.92} \\
\midrule
\textbf{Accuracy} & \multicolumn{3}{c}{0.86} & \multicolumn{3}{c}{0.87} & \multicolumn{3}{c}{0.87} \\
\bottomrule
\end{tabular}
}
\end{table}

BioBERT, which is pre-trained in biomedical corpora, is balanced in all categories. It outperforms BERT on four out of five classes and offers the best result on the \textit{Objective}, indicating its advantage in handling medically nuanced language.

SciBERT has the highest F1 score, with its robust performance in the \textit{Background} and \textit{Conclusions} being highlighted as its strength. The broader scientific domain is proposed to be responsible for supplying wider sentence structures typically seen in RCT abstracts (see Table~\ref{tab:per_class_metrics}).

Based on the results obtained by the models, BioBERT is the most consistent and well-balanced language model. SciBERT is appropriate for classifying scientific literature into different classes and achieving high accuracy, while BERT is weaker in terms of overall performance (as presented in Table~\ref{tab:model_recommendation}).

\begin{table}[H]
\centering
\caption{Model Recommendations based on Task Focus.}
\label{tab:model_recommendation}
\begin{tabular}{ll}
\toprule
\textbf{Use Case} & \textbf{Recommended Model} \\
Best overall balance & BioBERT \\
Scientific abstracts with varied language & SciBERT \\
Tasks dominated by methods class & BERT \\
\bottomrule
\end{tabular}
\end{table}

\subsubsection{Per-Category Performance}

\begin{figure}[h]
  \centering
  \includegraphics[width=0.7\linewidth]{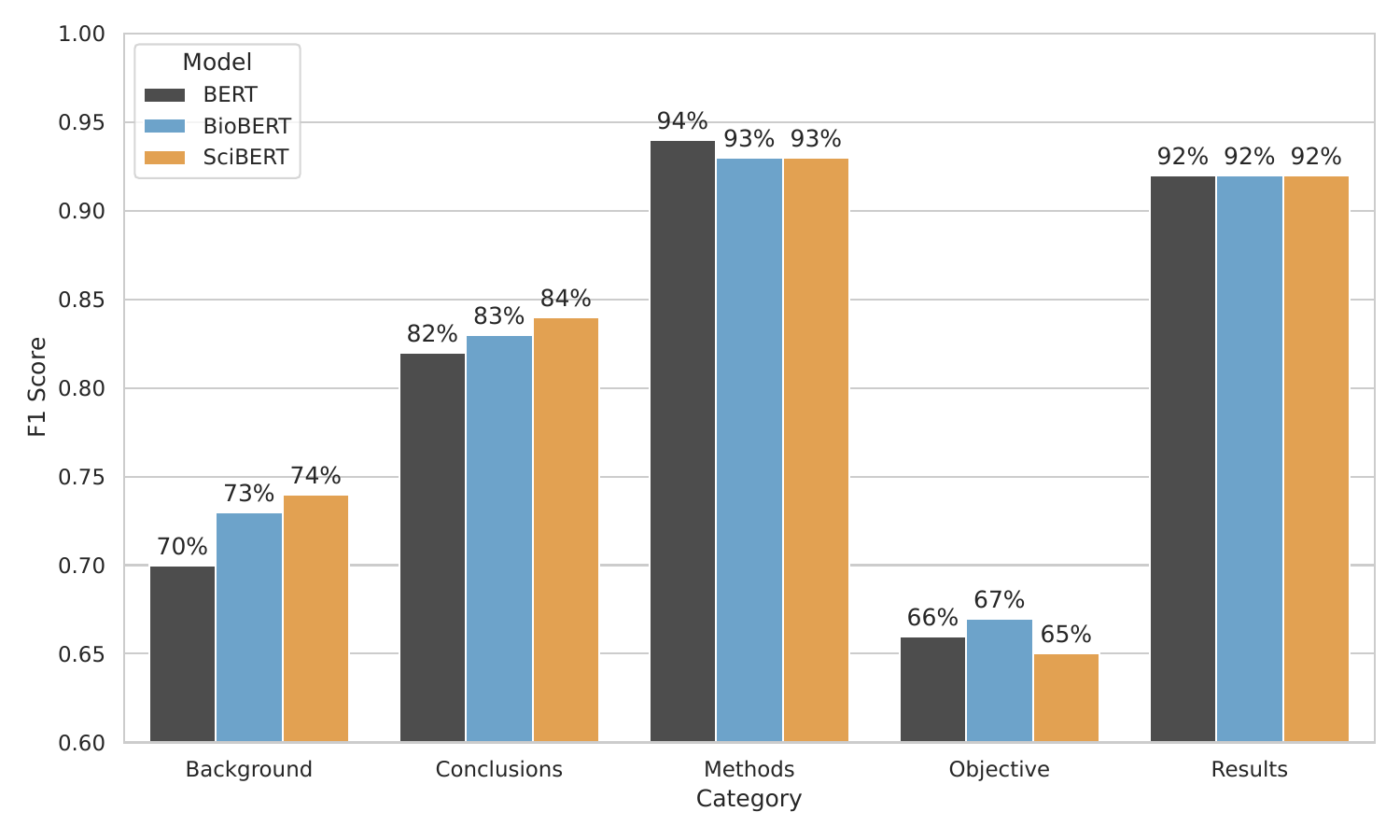}
  \caption{Per-category F1 score comparison.}
  \Description[Performance analysis per category]{This figure presents F1 score per category while utilizing BERT, SciBERT, and BioBERT.}
  \label{fig10}
\end{figure}

Table~\ref{tab:per_class_f1} provides the F1 score for each of the five discourse roles across the three models. SciBERT performs best on the \textit{Background} and \textit{Conclusions}, while BERT slightly outperforms others on the \textit{Methods}. BioBERT achieves the highest score for the \textit{Objective} category, which is consistently the most challenging class across all models (see Fig.~\ref{fig10}).

\begin{table}[H]
\centering
\caption{Summary of Per-Category F1 Score Comparison Across Models.}
\label{tab:per_class_f1}
\begin{tabular}{lcccc}
\toprule
\textbf{Class} & \textbf{BERT} & \textbf{BioBERT} & \textbf{SciBERT} & \textbf{Best Model} \\

Background & 0.70 & 0.73 & \textbf{0.74} & SciBERT \\
Conclusions & 0.82 & 0.83 & \textbf{0.84} & SciBERT \\
Methods & \textbf{0.94} & 0.93 & 0.93 & BERT \\
Objectives & 0.66 & \textbf{0.67} & 0.65 & BioBERT \\
Results & 0.92 & 0.92 & 0.92 & All Equal \\
\bottomrule
\end{tabular}
\end{table}

\section{Conclusion}
\label{conclusion}
In this SLR, we comprehensively investigated the utilization of PLMs for domain-specific text classification. Our study, encompassing 41 articles published between 2018 and January 2024, highlighted the significant advancements and persistent challenges in leveraging PLMs for specialized contexts. By categorizing existing research and presenting a taxonomy of techniques used, we presented the best-achieved performance of LLMs across various domains such as biomedical, computer science, finance, nuclear science, humanitarian, social science, and materials science. Our findings indicate that PLMs, particularly transformer-based models, offer substantial improvements in text classification tasks by effectively handling specialized vocabularies and unique grammatical structures. However, challenges such as imbalanced data distributions, high computational costs, and domain adaptation limitations persist. We explored and conducted experiments showing and visualizing that domain-specific PLMs achieve SOTA performance in their respective domains, but their efficacy is often contingent on extensive domain-specific training data and sophisticated fine-tuning strategies. We presented a comparative study to further underscore the variability in model performance across different domains, influenced by factors such as task complexity and data quality. We conclude that techniques like transfer learning, prompt-based learning, and activation fine-tuning show promise in enhancing model robustness and accuracy, yet they demand significant computational resources and expertise.

\section{Future Directions}
\label{sec:future_directions}
In this work, we highlighted advancements and existing challenges of utilizing PLMs for domain-specific text classification. To facilitate future research and improve the performance and applicability of PLMs in specialized contexts, the following research directions need to be addressed.

\noindent \emph{Enhanced Domain Adaptation Techniques:}
\begin{itemize}
    \item Development of more efficient methods for domain adaptation to handle specialized vocabulary and unique grammatical structures in different fields.
    \item Exploration of advanced transfer learning and fine-tuning strategies to improve model performance across various domains.
\end{itemize}
\emph{Optimization of Computational Resources:}
\begin{itemize}
    \item Research into reducing the computational cost of training and fine-tuning large language models, making them more accessible to smaller institutions and research teams.
    \item Investigation of lightweight model architectures that maintain high performance while requiring fewer resources.

\end{itemize}
\emph{Improved Data Utilization:}
\begin{itemize}
    \item Creation of larger, high-quality domain-specific datasets to support the training and evaluation of PLMs.
    \item Development of synthetic data generation techniques to supplement limited training data, ensuring model robustness and accuracy.

\end{itemize}
\emph{Ethical and Bias Mitigation Strategies:}
\begin{itemize}
    \item Addressing ethical concerns related to data privacy, bias, and transparency in model outputs.
    \item Implementation of techniques to mitigate biases in PLMs, particularly in sensitive domains such as finance and social sciences.

\end{itemize}
\emph{Advanced Prompt Engineering:}
\begin{itemize}
    \item Further refinement of prompt-based learning techniques, including the development of task-specific prompts to enhance model performance in zero-shot and few-shot learning scenarios.
    \item Exploration of novel prompt ensemble strategies and CoT reasoning to improve model outputs.

\end{itemize}
\emph{Comprehensive Benchmarking and Evaluation:}
\begin{itemize}
    \item Establishment of standardized benchmarks and evaluation metrics across different domains to facilitate consistent and comparative assessment of PLM performance.

\end{itemize}
By addressing these future directions, researchers can ensure that domain-specific PLMs continue to evolve and contribute significantly to various fields. This will ultimately lead to the development of more robust, efficient, and ethical NLP solutions for real-world applications.

\section{Limitations}
\label{sec:limitation}
This SLR offers valuable insights into the progress of domain-specific PLMs for text classification. However, several limitations are important to consider:
\begin{itemize}
    \item Risk of Bias: Despite utilizing AI-powered tools like Rayyan and ResearchRabbit for screening and selection, there remains a potential risk of bias due to the reliance on these tools for the initial selection of studies.
    \item Limited Data Range: The study only considered articles published between 2018 and January 2024, which might exclude relevant studies published outside this timeframe, thereby limiting the comprehensiveness of the review.
    \item Publication Bias: The strict inclusion criteria and the focus on specific keywords in titles, abstracts, or keywords may have led to the exclusion of relevant studies that did not explicitly mention the required terms, potentially introducing publication bias.
    
\end{itemize}

\section*{List of Abbreviations}
\null
{\footnotesize
\begin{tabular}{rl}
BART & Bidirectional and Auto-Regressive Transformers\\
BERT & Bidirectional Encoder Representations from Transformers\\
BLEU & BiLingual Evaluation Understudy\\
BLURB & Biomedical Language Understanding and Reasoning Benchmark\\
CARP & Clue And Reasoning Prompting\\
DL & Deep Learning\\
GRU & Gated Recurrent Unit\\
ICL & In-Context Learning\\
JRE & Joint Entity Recognition\\
LM & Language Model\\
LLM & Large Language Model\\
LLaMA & Large Language Model Meta AI\\
LSTM & Long Short Term Memory\\
METEOR & Metric for Evaluation of Translation with Explicit ORdering\\
MOF & Metal-Organic Frameworks\\
NER & Named Entity Recognition\\
NLI & Natural Language Inference\\
NLP & Natural Language Processing\\
NLU & Natural Language Understanding\\
NN & Neural Network\\
OPT & Open Pre-trained Transformer\\
PLM & Pre-trained Language Model\\
PRISMA & Preferred Reporting Items for Systematic Reviews and Meta-Analyses\\
PTM & Pre-trained Model\\
QuA & Question Answering\\
RE & Relation Extraction\\
REGEN & Retrieval-Enhanced Zero-Shot Data Generation\\
RNN & Recurrent Neural Network\\
ROUGE & Recall-Oriented Understudy for Gisting Evaluation\\
SLM & Small Language Model\\
SLR & Systematic Literature Review\\
SS & Semantic Similarity\\
SVM & Support Vector Machine\\
TF-IDF & Term Frequency-Inverse Document Frequency\\

\end{tabular}}

\section*{BERT Variants Stated in this SLR}
\null
{\footnotesize
\begin{tabular}{rl}
BioALBERT & Bio Adaptation of a Lite BERT \\
BioBART & Bio Bidirectional and Auto-Regressive Transformers\\
BioBERT & Bio BERT\\
DeBERTa & Decoding-enhanced BERT with disentangled attention\\
DistilBERT & Distlled BERT\\
FinBERT & Financial BERT\\
HumBERT & Humanitarian BERT\\
MatSciBERT & Material Scientific BERT\\
MediBioDeBERTa & Medical Bio DeBERTa\\
NukeBERT & Nuclear BERT\\
RadBERT & Radiology BERT\\
RoBERTa & Robustly Optimized BERT Approach\\
SciBERT & Scientific BERT\\
SciDeBERTa & Scientific DeBERTa\\
SSciBERT & Social SciBERT\\

\end{tabular}}

\begin{acks}
The authors are grateful to the members of the Applied Machine Learning Research Group of Óbuda University John von Neumann Faculty of Informatics for constructive comments and suggestions. The authors would also like to acknowledge the support of the Doctoral School of Applied Informatics and Applied Mathematics of Óbuda University.
\end{acks}

\bibliographystyle{ACM-Reference-Format}
\bibliography{sample-base}

\appendix

\end{document}